\documentclass{article}

    \usepackage[preprint]{neurips_2021}

\usepackage[utf8]{inputenc} % allow utf-8 input
\usepackage[T1]{fontenc}    % use 8-bit T1 fonts
\usepackage{hyperref}       % hyperlinks
\usepackage{url}            % simple URL typesetting
\usepackage{booktabs}       % professional-quality tables
\usepackage{amsfonts}       % blackboard math symbols
\usepackage{nicefrac}       % compact symbols for 1/2, etc.
\usepackage{microtype}      % microtypography
\usepackage{xcolor}         % colors
\usepackage{comment} 

\usepackage{microtype}
\usepackage[pdftex]{graphicx}
\usepackage{subfigure}
\usepackage{booktabs} % for professional tables
\usepackage{algorithm}
\usepackage{algorithmic}
\usepackage{amsmath, amsthm, amssymb}

\usepackage{color}
\usepackage{comment}
\usepackage{multirow}
\usepackage[font=small,labelfont=bf, tableposition=top]{caption}

\newcommand{\blue}[1]{{\color{black} #1}}

\title{Learning by Examples Based on \\Multi-level Optimization}

\begin{document}

% The \author macro works with any number of authors. There are two commands
% used to separate the names and addresses of multiple authors: \And and \AND.
%
% Using \And between authors leaves it to LaTeX to determine where to break the
% lines. Using \AND forces a line break at that point. So, if LaTeX puts 3 of 4
% authors names on the first line, and the last on the second line, try using
% \AND instead of \And before the third author name.

\author{%
  Shentong Mo \\
  Carnegie Mellon University\\
  Pittsburgh, PA 15213 \\
  \texttt{shentonm@andrew.cmu.edu} \\
   \And
   Pengtao Xie \\
  University of California, San Diego\\
   La Jolla, CA, 92093 \\
  \texttt{pengtaoxie2008@gmail.com}
}\maketitle

\begin{abstract}

 Learning by examples, which learns to solve a new problem by looking into how similar problems are solved, is an effective learning method in human learning. 
 When a student learns a new topic, he/she finds out exemplar topics that are similar to this new topic and studies the exemplar topics to deepen the understanding of the new topic. 
 We aim to investigate whether this powerful learning skill can be borrowed from humans to improve machine learning as well. 
 In this work, we propose a novel learning approach called Learning By Examples (\textbf{LBE}). 
 Our approach automatically retrieves a set of training examples that are similar to query examples and predicts labels for query examples by using class labels of the retrieved examples. 
 We propose a three-level optimization framework to formulate LBE which involves three stages of learning: learning a Siamese network to retrieve similar examples; learning a matching network to make predictions on query examples by leveraging class labels of retrieved similar examples; 
 learning the ``ground-truth'' similarities between training examples by minimizing the validation loss. 
 We develop an efficient algorithm to solve the LBE problem and conduct extensive experiments on various benchmarks where the results demonstrate the effectiveness of our method on both supervised and few-shot learning.

\end{abstract}
% \vspace{-1em}
\section{Introduction}
% \vspace{-0.5em}
Learning by examples is a broadly used technique in human learning. For example, in course learning, given a new topic for students, students will find out exemplar topics that are similar to this new topic and study the exemplar topics to deepen the understanding of the new topic. The study in~\blue{\cite{frederick1978lbe,frederick1976learning,patrick1975learning}} shows that learning by examples is a powerful tool for helping people learn new things.

Inspired by this examples-driven learning technique of humans, we are interested in
investigating whether this methodology can be helpful for improving the abilities of machine learning as well.
We propose a novel learning framework called learning by examples (LBE). In this framework, the model is trained to retrieve a set of training examples that are similar to the query examples and predict labels for query examples by using the class labels of the retrieved examples. The model consists of a Siamese network and a matching network.  The Siamese network is trained to retrieve similar examples and the matching network is trained to make predictions on query examples by leveraging retrieved similar examples. 
Our framework is formulated as a three-level optimization problem that involves three learning stages. 
In the first learning stage, we train a Siamese network $T$ by minimizing the cross-entropy loss between predicted similarity by $T$ and the ground-truth similarity.

In this stage, the ground-truth similarities are fixed. 
They will be updated at a later stage.
The optimal solution $T^*(A)$ is a function of $A$ since $T^*(A)$ is a function of the cross-entropy loss and the cross-entropy loss is a function of $A$. 
In the second stage, we use the Siamese network $T^*(A)$ trained in the first stage to retrieve a subset of training examples $B$ similar to each query example. 
Then we train the matching network $S$ to predict labels for query examples using the labels of retrieved examples $B$.
Note that the optimal solution $S^*(T^*(A))$ is a function of $T^*(A)$ since $S^*(T^*(A))$ is a function of the cross entropy loss and the cross entropy loss is a function of $T^*(A)$. 
In the third stage, we apply $S^*(T^*(A))$ to make predictions on the validation set and update $A$ by minimizing the validation loss. 
The three stages are performed jointly end-to-end, where different stages influence each other. 
Experiments on various benchmarks demonstrate the effectiveness of our
method.

\vspace{-0.8em}

\begin{figure}[h]
    \centering
    \includegraphics[width=0.6\textwidth]{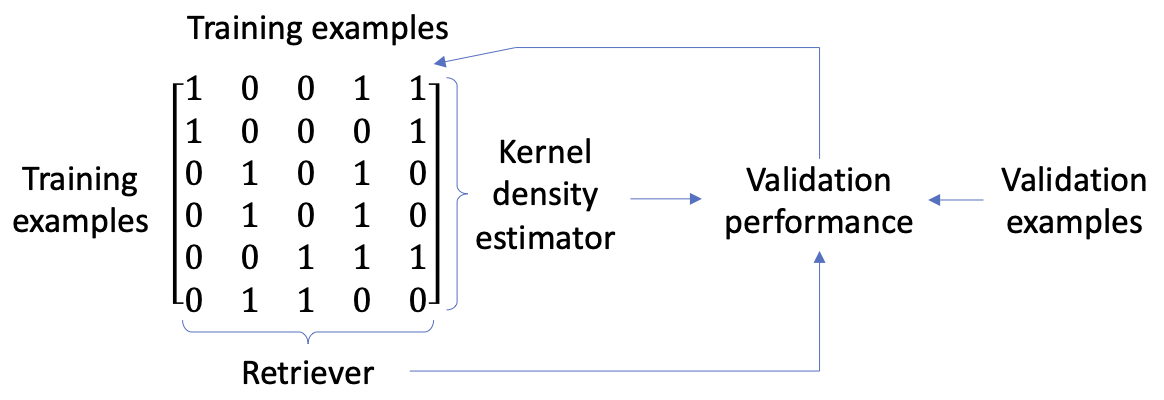}
    \caption{Illustration of learning by examples. The Retriever (a Siamese network $T$) and the Kernel density estimator (a matching network $S$) are used to make predictions for Validation examples. 
    The similarity matrix $A$ of training examples is updated by minimizing corresponding losses of Validation performance.
    }
    \label{fig:illus}
\end{figure}

The major contributions of this paper are summarized as follows:
\begin{itemize}
\item Inspired by the examples-driven learning technique of humans, we propose a novel
machine learning approach called learning by examples (LBE). Our approach automatically retrieves a set of training examples that are similar to query examples and predicts labels for query examples by using the class labels of the retrieved ones.
\item We propose a multi-level optimization framework to formulate LBE which involves three stages of learning: learning to retrieve similar examples by a Siamese network; 
 learning to make predictions using retrieved examples by a matching network; 
 learning the ``ground-truth'' similarity matrix between training examples by minimizing the validation loss.
 \item We develop an efficient optimization algorithm to solve the LBE problem.
\item Experiments on various benchmarks demonstrate the effectiveness of our method on supervised and few-shot learning.
\end{itemize}

The rest of the paper is organized as follows. Section~\blue{\ref{method}} and~\blue{\ref{experiments}} present the method and
experiments respectively. Section~\blue{\ref{related_work}} reviews related works. Section~\blue{\ref{conclusion}} concludes the paper.
\section{Method}\label{method}\label{sec:method}
% \vspace{-0.5em}

In this section, we propose a framework called learning by examples (LBE) and develop an optimization algorithm for solving the LBE problem in Figure~\blue{\ref{fig: illus_opt}}. The solid arrows denote the process of making predictions and calculating losses. The dotted arrows denote the process of updating learnable parameters by minimizing corresponding losses. To easy reading, we summarize the notations in Table 1.
% ~\blue{\ref{tab:notation}}.
% \input{SECTIONS/30_Method/notation}
\begin{minipage}{\linewidth}
\begin{minipage}{.3\linewidth}
	\centering
		%\fbox{\rule{0pt}{2in} \rule{0.8\linewidth}{0pt}}
	\includegraphics[width=0.8\linewidth]{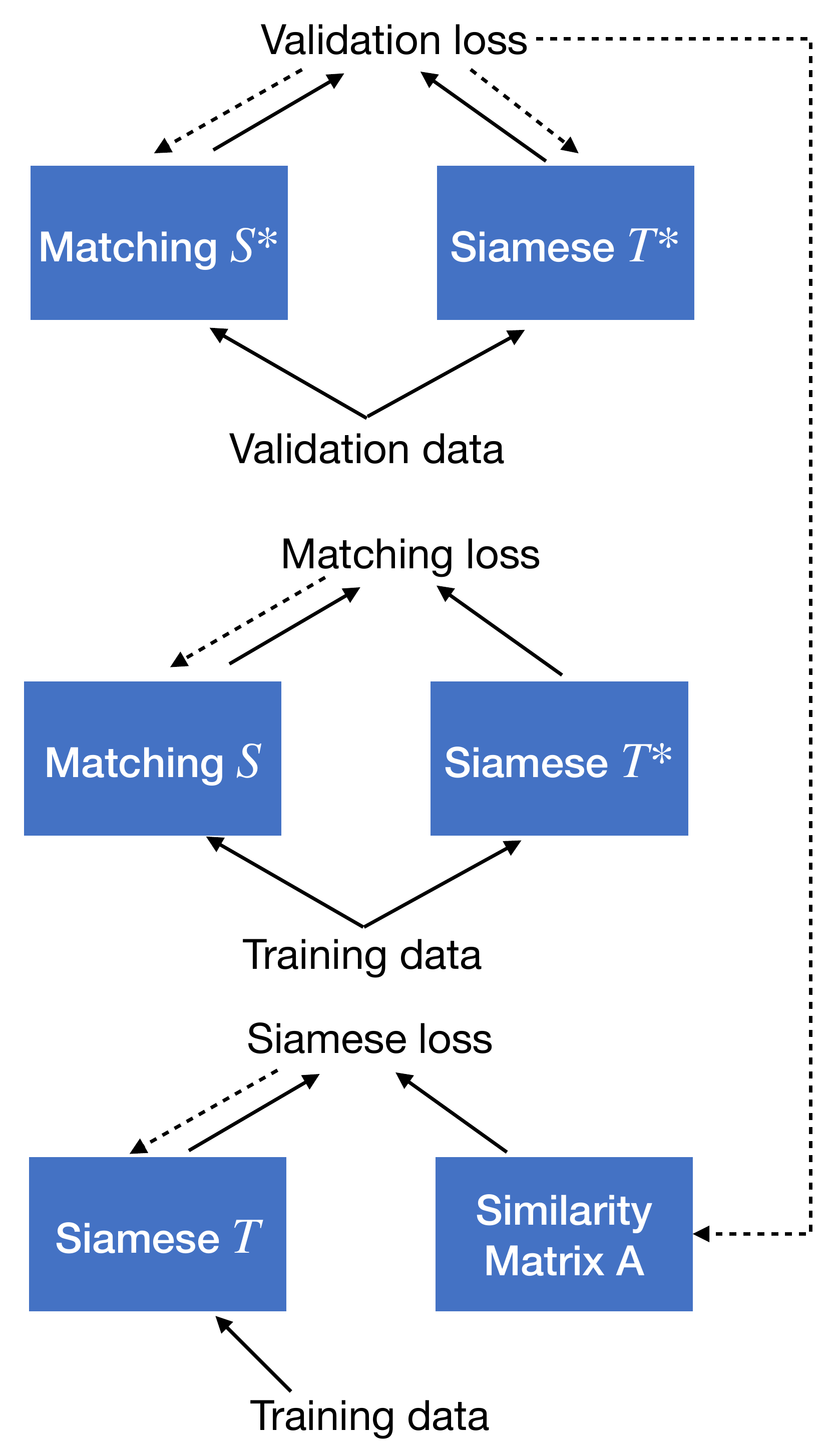}
	\captionof{figure}{Learning by examples. }
	\label{fig: illus_opt}
\end{minipage}
\hfill
\begin{minipage}{.7\linewidth}
	%\normalem
	\label{tab:notation}
	\renewcommand\tabcolsep{6.0pt}
	\centering
	\captionof{table}{Notations in learning by examples.}
	\scalebox{0.9}{
		\begin{tabular}{l|l}
			\toprule
			Notation & Meaning  \\ 	
			\midrule
			$N$ & the number of training examples \\
			$M$ & the number of validation examples \\
			$x_i, x_j$ & two training examples $i$ and $j$ \\
            $y_i, y_j$ & the groundtruth label of $x_i$ and $x_j$ \\
			$u_i$ & a validation example $i$ \\
			$t_i$ & the groundtruth label of $u_i$ \\
			$A$ & the similarity matrix between training examples \\
			$A_{i,j}$ & the similarity matrix between training examples $i$ and $j$ \\
			$T$ & the parameters of Siamese network \\
			$S$ & the parameters of matching network \\
			$D^{(\text{tr})}$ & training dataset \\
			$D^{(\text{val})}$ & validation dataset \\
			
			\bottomrule
			\end{tabular}}
		    
\end{minipage}
\end{minipage}

% \begin{table}
% 	\begin{minipage}{0.4\linewidth}
% 		\caption{Caption }
% 		\label{tab:le}
% 		\centering
% 		\resizebox{\textwidth}{!}{%
% 		\begin{tabular}[width=\linewidth]{@{}lll@{}}
%     \toprule
%     \textbf{Student} & \textbf{week} & \textbf{grade} \\ 
%     \midrule
%     Ada Lovelace & 2 & A \\
% 			Linus Thorvalds & 8 & A \\
% 			Bruce Willis & 12 & F \\
% 			Richard Stallman & 10 & B \\
% 			Grace Hopper & 12 & A \\
% 			Alan Turing & 8 & C \\
% 			Bill Gates & 6 & D \\
% 			Steve Jobs & 4 & E \\
%     \bottomrule
%     \end{tabular}}
% 	\end{minipage}\hfill
% 	\begin{minipage}{0.6\linewidth}
% 		\centering
% 		\captionof{figure}{Caption_2}
% 		\label{figu:re}
% 		\includegraphics[width=0.8\textwidth]{SECTIONS/30_Method/illustration.pdf}
% 	\end{minipage}
% \end{table}

% \setlength{\belowdisplayskip}{0.7pt} \setlength{\belowdisplayshortskip}{0.7pt}
% \setlength{\abovedisplayskip}{0.7pt} \setlength{\abovedisplayshortskip}{0.7pt}

\subsection{Learning by Examples}
% \vspace{-0.5em}
Learning from examples is a broadly used human learning technique. When a student learns a new topic, he/she finds out exemplar topics that are similar to this new topic and studies the exemplar topics to deepen the understanding of the new topic. We aim to leverage this human learning skill to help with machine learning. Given a query example, we first retrieve a set of training examples that are similar to the query. Then we use the class labels of the retrieved examples to predict the label of the query.

Without loss of generality, we assume the end task is image classification while noting that our framework can be applied to other tasks as well. 
Let $N$ be the number of training examples. Let $A$ be a $N \times N$ learnable matrix, where $A_{ij} \in [0, 1]$ indicates the similarity between training example $i$ and $j$. A larger $A_{ij}$ indicates more similarity. Let $f(x_i,x_j;T)$ be a Siamese network with network weights $T$. $f(x_i, x_j; T)$ takes two images $x_i$ and $x_j$ as inputs and outputs a probability indicating how similar $x_i$ and $x_j$ are. Our framework involves three inter-dependent learning stages. In the first stage, we train $T$ by solving the following optimization problem:
\begin{equation}
    T^*(A) = \min_{T}\sum_{i=1}^{N-1}\sum_{j=i+1}^{N} -A_{i,j}\log f(x_i,x_j;T)
\end{equation}
where $A_{ij}$ is treated as the ``ground-truth'' similarity between example $i$ and $j$. The Siamese network is trained to match the ``ground-truth''. In the second stage, for each training example $x_i$, we use the Siamese network trained in the first stage to retrieve a subset of training examples $B$ that are similar to $x_i$. Then we train a matching network parameterized by $S$. The matching network predicts the label $\hat{y}_{i}$ for $x_i$ in the following way:
\begin{equation}\label{eq:SecondEquation}
    \hat{y}_{i} = \sum_{x_j\in B} c(x_i, x_j; S)y_j
\end{equation}

where $c(x_i, x_j, S)$ calculates the similarity between $x_i$ and $x_j$. $y_j$ is the groundtruth label of $x_j$. Eq.(\ref{eq:SecondEquation}) can be relaxed to 
\begin{equation}
    \hat{y}_{i} = \sum_{j=1}^N f(x_i, x_j; T^*(A))c(x_i, x_j, S)y_j
\end{equation}

We learn $S$ by solving the following optimization problem:
\begin{equation}
    S^*(T^*(A))= \min_{S}\sum_{i=1}^N -y_i\log (\sum_{j=1}^N f(x_i, x_j; T^*(A))c(x_i, x_j, S)y_j)
\end{equation}

where $y_i$ is the ground-truth label of $x_i$. Given the trained Siamese network and matching network, we perform prediction on the validation examples and get the following validation loss
\begin{equation}
    \sum_{i=1}^M -t_i\log (\sum_{j=1}^N f(u_i, x_j; T^*(A))c(u_i, x_j, S^*(T^*(A)))y_j)
\end{equation}

where $M$ is the number of validation examples, $u_i$ is a validation example and $t_i$ is its class label. We learn the matrix $A$ by minimizing this loss. Putting these pieces together, we get the following overall formulation. 
\begin{equation}\label{eq:overall}
\begin{aligned}
\min_{A}\ & \sum_{i=1}^{M} -t_{i}\log (\sum_{j=1}^{N}f(u_i,x_j;T^*(A))c(u_i,x_j;S^*(T^*(A)))y_j) \\
\text{s.t.}\ & S^*(T^*(A)) = \min_{S}\sum_{i=1}^{N}-y_i\log (\sum_{j=1}^{N}f(x_i,x_j;T^*(A))c(u_i,x_j;S)y_j) \\
& T^*(A) = \min_{T}\sum_{i=1}^{N-1}\sum_{j=i+1}^{N} -A_{i,j}\log f(x_i,x_j;T)
\end{aligned}
\end{equation}

Note that there are multiple notations of similarity in this formulation. The Siamese network calculates similarities between the query and all training examples for retrieving similar training examples. The matching network calculates similarities between query and retrieved examples to predict the class label for the query. The similarity in $A$ is the ``ground-truth'' similarity between training examples.  

% \input{SECTIONS/30_Method/fig_illustration}
% \vspace{-1.5em}
\subsection{Optimization Algorithm}
\label{optimization_algorithm}
% \vspace{-0.5em}

In this section, we derive
an optimization algorithm for solving the LBE problem defined in Eq.(\blue{\ref{eq:overall}}). 
\begin{comment}
\begin{equation}\label{eq:overall}
\begin{aligned}
\min_{A}\ & \sum_{i=1}^{M} -t_{i}\log (\sum_{j=1}^{N}f(u_i,x_j;T^*(A))c(u_i,x_j;S^*(T^*(A)))y_j) \\
s.t.\ & S^*(T^*(A)) = \min_{S}\sum_{i=1}^{N}-y_i\log (\sum_{j=1}^{N}f(x_i,x_j;T^*(A))c(u_i,x_j;S)y_j) \\
& T^*(A) = \min_{T}\sum_{i=1}^{N-1}\sum_{j=i+1}^{N} -A_{i,j}\log f(x_i,x_j;T)
\end{aligned}
\end{equation}
\end{comment}
To explain how to solve the optimization problem, we denote the above formulation simply as

\vspace{-1em}
\begin{equation}\label{eq:simple}
\begin{aligned}
\min_{A}\ &L(T^*(A), S^*(T^*(A)), D^{(\text{val})}) \\
s.t.\ &S^*(T^*(A))=\min_{S}L(T^*(A), S, D^{(\text{tr})}) \\
& T^*(A) = \min_{T}L(A, T, D^{(\text{tr})})
\end{aligned}
\end{equation}
where $D^{(\text{tr})}$ and $D^{(\text{val})}$ to denote the training data and validation data.
\begin{comment}
\begin{equation}\label{eq:simple2}
\setlength{\abovedisplayskip}{3pt}
\setlength{\belowdisplayskip}{3pt}
\begin{aligned}
L(T^*(A), S^*(T^*(A)), D^{(\text{val})}) &= \sum_{i=1}^{M} -t_{i}\log (\sum_{j=1}^{N}f(u_i,x_j;T^*(A))c(u_i,x_j;S^*(T^*(A)))y_j)\\
L(T^*(A), S, D^{(\text{tr})}) &= \sum_{i=1}^{N}-y_i\log (\sum_{j=1}^{N}f(x_i,x_j;T^*(A))c(u_i,x_j;S)y_j) \\
L(A, T, D^{(\text{tr})}) &= \sum_{i=1}^{N-1}\sum_{j=i+1}^{N} -A_{i,j}\log f(x_i,x_j;T)
\end{aligned}
\end{equation}
\end{comment}
Inspired by \blue{\cite{liu2018darts}}, we approximate $T^*(A)$ using one-step gradient descent update of $T$ with respect to $L(A, T, D^{(\text{tr})})$, then plug in $T^*(A)$ into $L(T^*(A), S, D^{(\text{tr})})$ to approximate $S^*(T^*(A))$ using one-step
gradient descent update of $S$ with respect to its objective. Then we plug in these two approximations into $L(T^*(A), S^*(T^*(A)), D^{(\text{val})})$ and perform gradient-descent update of $A$ with respect to this approximated objective. In the sequel, we use $\nabla^{2}_{Y,X} f(X,Y)$ to denote $\dfrac{\partial{f(X,Y)}}{\partial{X}\partial{Y}}$.

Approximating $T^*(A)$ using $T^\prime = T - \xi_{T}\nabla_{T}L(A, T, D^{(\text{tr})})$ where $\xi_{T}$ is a learning rate, we can calculate the approximated gradient of $L(T^*(A), S, D^{(\text{tr})})$ w.r.t $S$ as:
\begin{equation}\label{eq:simple4}
\begin{aligned}
&\nabla_{S}L(T^*(A), S, D^{(\text{tr})}) \approx \nabla_{S}L(T^\prime, S, D^{(\text{tr})})\\
\end{aligned}
\end{equation}
Then we can approximate $S^*(T^*(A))$ using $S^\prime = S - \xi_{S}\nabla_{S}L(T^\prime, S, D^{(\text{tr})})$ where $\xi_{S}$ is also a learning rate. 
We can approximate $T^*(A)$ and $S^*(T^*(A))$ using the following one-step gradient descent update of $T$ and $S$ respectively:
\begin{equation}\label{eq:simple5}
\begin{aligned}
T^\prime = T - \xi_{T}\nabla_{T}L(A, T, D^{(\text{tr})}), \quad S^\prime = S - \xi_{S}\nabla_{S}L(T^\prime, S, D^{(\text{tr})})
\end{aligned}
\end{equation}
% and 
% \begin{equation}\label{eq:simple6}
% \begin{aligned}
% &S^\prime = S - \xi_{S}\nabla_{S}L(T^\prime, S, D^{(\text{tr})})
% \end{aligned}
% \end{equation}
where $\xi_{T}$ and $\xi_{S}$ are learning rates. 
\iffalse
\fi
Plugging in these approximations into the objective function $L(T^*(A), S^*(T^*(A)), D^{(\text{val})})$, we can learn $A$ by minimizing the  objective $L(T^\prime, S^\prime, D^{(\text{val})})$ using gradient methods. The derivative of this objective with respect to $A$ can be calculated as:
\begin{equation}\label{eq:simple7}
\begin{aligned}
&\nabla_{A}L(T^\prime, S^\prime, D^{(\text{val})}) = \dfrac{\partial{T^\prime}}{\partial{A}}\nabla_{T^\prime}L(T^\prime, S^\prime, D^{(\text{val})}) + \dfrac{\partial{S^\prime}}{\partial{A}}\nabla_{S^\prime}L(T^\prime, S^\prime, D^{(\text{val})})
\end{aligned}
\end{equation}
where
\begin{equation}\label{eq:simple8}
\begin{aligned}
\dfrac{\partial{T^\prime}}{\partial{A}} = -\xi_{T}\nabla^2_{A,T}L(A, T, D^{(\text{tr})}), \quad \dfrac{\partial{S^\prime}}{\partial{A}} = -\xi_{S}\dfrac{\partial{T^\prime}}{\partial{A}}\nabla^2_{T^\prime,S}L(T^\prime, S, D^{(\text{tr})})
\end{aligned}
\end{equation}

$\nabla^2_{A,T}L(A, T, D^{(\text{tr})})\nabla_{T^\prime}L(T^\prime, S^\prime, D^{(\text{val})})$ involves expensive matrix-vector product. We can reduce the computational complexity by a finite difference approximation:
\begin{equation}\label{eq:simple9}
\begin{aligned}
\nabla^2_{A,T}&L(A, T, D^{(\text{tr})})\nabla_{T^\prime}L(T^\prime, S^\prime, D^{(\text{val})}) \approx \\ 
& \dfrac{1}{2\alpha_{T}}\Big(\nabla_{A}L(A, T^{+}, D^{(\text{tr})}) - \nabla_{A}L(A, T^{-}, D^{(\text{tr})})\Big) 
\end{aligned}
\end{equation}
where $T^{\pm}=T\pm\alpha_{T}\nabla_{T^\prime}L(T^\prime, S^\prime, D^{(\text{val})})$, and $\alpha_{T}$ is a small scalar that equals to $$0.01/\lVert{\nabla_{T^\prime}L(T^\prime, S^\prime, D^{(\text{val})})\rVert}_2.$$ 

Similar to Eq.(\blue{\ref{eq:simple9}}), using finite difference approximation to calculate $$\nabla^2_{T^\prime,S}L(T^\prime, S, D^{(\text{tr})}\nabla_{S^\prime}L(T^\prime, S^\prime, D^{(\text{val})}),$$ we have:
\begin{equation}\label{eq:simple10}
\begin{aligned}
\nabla^2_{T^\prime,S}&L(T^\prime, S, D^{(\text{tr})})\nabla_{S^\prime}L(T^\prime, S^\prime, D^{(\text{val})}) \approx \\ &\dfrac{1}{2\alpha_{S}}\Big(\nabla_{T^\prime}L(T^\prime, S^{+}, D^{(\text{tr})}) - \nabla_{T^\prime}L(T^\prime, S^{-}, D^{(\text{tr})})\Big) 
\end{aligned}
\end{equation}
where $S^{\pm}=S\pm\alpha_{S}\nabla_{S^\prime}L(T^\prime, S^\prime, D^{(\text{val})})$, and $\alpha_{S}$ is a small scalar that equals to $$0.01/\lVert{\nabla_{S^\prime}L(T^\prime, S^\prime, D^{(\text{val})})\rVert}_2.$$ The overall algorithm for solving the LBE problem is summarized in Algorithm \blue{\ref{alg:alg_optimization}}.

\begin{algorithm}[h]
   \caption{Optimization algorithm for learning by examples}
   \label{alg:alg_optimization}
\begin{algorithmic}
   \WHILE{not converge}
        \STATE 1. Update the matrix $A$ by descending the gradient calculated in Eq.(\blue{\ref{eq:simple7}}-\blue{\ref{eq:simple10}}) \; \\
        \STATE 2. Update the the matching network parameters $S$ by descending the gradient in Eq.(\blue{\ref{eq:simple5}}) \;\\
        \STATE 3. Update the weights $T$ of the Siamese network by descending the gradient in Eq.(\blue{\ref{eq:simple5}})\; \\
   \ENDWHILE
\end{algorithmic}
\end{algorithm}

\subsection{Fast Retrieval \& Matching}
\label{fast_retrieval}
% \vspace{-0.5em}

When the training dataset is large, performing retrieval on a large number of training examples during test time is computationally inefficient. 
To address this problem, we design the retrieval network $T$ in a way that facilitates fast retrieval.
Given two images $x_i$ and $x_j$, we first encode them using a CNN to get their encodings $\mathbf{e}_i$ and $\mathbf{e}_j$. 
In the encoding vector, the value of each dimension is in [0, 1] (after applying sigmoid).
Then we transform these two encoding vectors into binary vectors $\mathbf{h}_i$ and $\mathbf{h}_j$, where elements in binary vectors are either 0 or 1. 
The transformation is as follows:
for each dimension in an encoding vector where the value of this dimension is $p$, we define a Bernoulli distribution parameterized by $p$; 
then we sample a binary variable from this Bernoulli distribution. 
Given these two binary vectors $\mathbf{h}_i$ and $\mathbf{h}_j$, we calculate the similarity of the two images by taking the Hamming distance between $\mathbf{h}_i$ and $\mathbf{h}_j$.
The similarity is denoted by $f(x_i, x_j;T)$. 
The rest of Eq.(~\blue{\ref{eq:overall}}) remains the same.
The binary vectors are not end-to-end differentiable. To address this problem, we use the Gumbel-softmax~\blue{\cite{jang2017gumbel}} trick.
Hamming distance between binary vectors can be calculated very efficiently, which facilitates fast retrieval. 
There are three variants of the vanilla LBE model for fast retrieval.
(1) LBE-fast-$T$, where we apply hash codes for $T$ and continuous representations for $S$; 
(2) LBE-fast-$S$, where we apply hash codes for $S$ and continuous representations for $T$; 
(3) LBE-fast-$TS$, where we apply hash codes for both $T$ and $S$.

% \vspace{-1em}
\section{Experiments}\label{experiments}
% \vspace{-0.5em}
\subsection{Datasets}\label{datasets}
% \vspace{-0.5em}
We used four datasets in the experiments: CIFAR-10, CIFAR-100, ImageNet~\blue{\cite{deng2009imagenet}}, and Tiny-ImageNet~\blue{\cite{wu2017tiny}}. 
The CIFAR-10 dataset contains 50K training images and 10K testing images, from 10 classes (the number of images in each class is equal). We split the original 50K training set into a new 45K training set and a 5K validation set.
In the sequel, when we mention “training set”, it always refers to the new 45K training set. 
The CIFAR-100 dataset contains 50K training images and 10K testing images, from 100 classes (the number of images in each class is equal). 
Similar to CIFAR-10, the 50K training images are split into a 45K training set and 5K validation set. 
The usage of the new training set and validation set is the same as that for CIFAR-10. 
The ImageNet dataset contains a training set of 1.2M images and a validation set of 50K images, from 1000 object classes. The validation set is used as a test set for architecture evaluation.
The Tiny-ImageNet dataset is a modified subset of the original ImageNet dataset, where there are 200 different classes instead of 1000 classes of the ImageNet dataset, with
100K training examples and 10K
validation examples. 
The usage of the validation set is the same as that for the ImageNet dataset.

For few-shot experiments, we use Omniglot~\blue{\cite{brenden2011one}}, miniImageNet~\blue{\cite{deng2009imagenet}} and tiredImageNet~\blue{\cite{deng2009imagenet}}. We closely follow previous methods~\blue{\cite{vinyals2016matching}} and compare our LBE framework against strong baselines. 
All of our experiments revolve around the same basic task: an $N$-way
$k$-shot learning task. 
During the meta-training stage, we use $N$ classes and $k$ samples per class in the support set. Then we predict the label for the tested batch set which has the same classes as the support set.
In our LBE framework, we take this support set as the training set $D^{\text{(tr)}}$ and the batch set as the validation set $D^{\text{(val)}}$. Then we do the three-level optimization task in the LBE framework.
During the meta-test stage, we apply the same setting as the training stage and provide a support set ($N$ classes and $k$ samples per class) from unseen classes. 
We compared a number of alternative models, as baselines, to our LBE framework.
% \vspace{-0.5em}
\subsection{Baselines}\label{baselines}
% \vspace{-0.5em}
For experiments on CIFAR-10, CIFAR-100, ImageNet~\blue{\cite{deng2009imagenet}}, and Tiny-ImageNet~\blue{\cite{wu2017tiny}}, we compare with the supervised learning method as our baseline to validate the rationality of learning by examples. We also implement state-of-the-art ResNet~\blue{\cite{he2016resnet}} series models (ResNet-18, ResNet-34, ResNet-50, ResNet-101, ResNet-152, ResNeXt-50, ResNeXt-101, wide ResNet-50, wide ResNet-101) as the backbone for comprehensive comparison.
In order to prove the effectiveness of learning by examples on the few-shot setting, we also apply previous few-shot learning methods~\blue{\cite{vinyals2016matching,ravi2017optimization,finn2017modelagnostic,triantafillou2017fewshot,garcia2018fewshot,mishra2018a,oreshkin2018tadam,lee2018gradient,wang2019simpleshot,jiang2019learning,chen2019a}} as our baselines for a comprehensive comparison.

% \vspace{-0.5em}
\subsection{Experiment Settings}\label{exp_settings}
% \vspace{-0.5em}
The Siamese and matching network weights are optimized using SGD with a momentum of 0.9 and a weight decay of 3e-4. The initial learning rate is set to 0.01 with a cosine decay scheduler. The similarity matrix is optimized with the Adam~\blue{\cite{kingma2015adam}} optimizer with a learning rate of 1e-3 and a weight decay of 1e-3. The network is trained for 1000 epochs with a batch size of 32 (for both CIFAR-10 and CIFAR-100).  The experiments are performed on a single Tesla v100. For ImageNet and Tiny-ImageNet, the network is trained for 1000 epochs with a batch size of 64 on one Tesla v100 GPU. Each experiment on LBE is repeated ten times with the random seed to be from 1 to 10. We report the mean and standard deviation of results obtained from 10 runs using different random seeds.
% \vspace{-2.0em}
\subsection{Results}\label{exp_results}
% \vspace{-0.5em}
\subsubsection{Supervised Settings}
\begin{table}[!htb]
\setlength{\abovecaptionskip}{-0em}
\setlength{\belowcaptionskip}{-1em}
\begin{minipage}{.48\linewidth}
    \centering

    \caption{Comparison results on the \textit{CIFAR-10}.}
    \label{tab: exp_c10}

	%\normalem
	\renewcommand\tabcolsep{6.0pt}
	%\medskip
	\scalebox{0.6}{
		\begin{tabular}{l|c|c|c|c}
			\toprule
			\multicolumn{1}{l|}{\multirow{2}{*}{Backbone}} & \multicolumn{2}{c|}{Supervised} & \multicolumn{2}{c}{LBE}                     \\ \cline{2-5}  
			& top-1 & top-5 & top-1 & top-5 \\ 	\midrule
			
			ResNet-18 & 92.95$\pm$0.11 & 99.25$\pm$0.08 & 92.03$\pm$0.06 & 98.83$\pm$0.03 \\
			ResNet-34 & 93.23$\pm$0.12 & 99.43$\pm$0.07 & 92.31$\pm$0.06 & 98.96$\pm$0.03 \\
			ResNet-50 & 93.41$\pm$0.13 & 99.54$\pm$0.05 & 92.52$\pm$0.05 & 99.11$\pm$0.03 \\
			ResNet-101 & 93.69$\pm$0.11 & 99.61$\pm$0.04 & 92.83$\pm$0.05 & 99.22$\pm$0.02  \\
			ResNet-152 & 93.62$\pm$0.11 & 99.58$\pm$0.02 & 92.79$\pm$0.04 & 99.19$\pm$0.01 \\
			ResNeXt-50 & 96.21$\pm$0.13 & 99.82$\pm$0.03 & 95.47$\pm$0.06 & 99.53$\pm$0.02 \\
			ResNeXt101 & 96.36$\pm$0.09 & 99.85$\pm$0.02 & 95.66$\pm$0.05 & 99.57$\pm$0.02 \\
			wide\_ResNet50  & 95.83$\pm$0.13 & 99.76$\pm$0.03 & 95.02$\pm$0.06 & 99.33$\pm$0.02 \\
			wide\_ResNet101 & 96.11$\pm$0.12 & 99.79$\pm$0.05 & 95.52$\pm$0.06 & 99.45$\pm$0.02 \\
			\bottomrule
			\end{tabular}}
\end{minipage}\hfill
\begin{minipage}{.5\linewidth}
    \centering

    \caption{Comparison results on the \textit{CIFAR-100}.}
    \label{tab: exp_c100}

    %\medskip
	\scalebox{0.6}{
		\begin{tabular}{l|c|c|c|c}
			\toprule
			\multicolumn{1}{l|}{\multirow{2}{*}{Backbone}} & \multicolumn{2}{c|}{Supervised} & \multicolumn{2}{c}{LBE}                     \\ \cline{2-5}  
			& top-1 & top-5 & top-1 & top-5 \\ 	\midrule
			ResNet-18 & 75.61$\pm$0.11 & 93.05$\pm$0.05 &  73.55$\pm$0.07 & 92.27$\pm$0.03 \\
			ResNet-34 & 76.76$\pm$0.13 & 93.37$\pm$0.04 & 75.36$\pm$0.07 & 92.62$\pm$0.03 \\
			ResNet-50 & 77.39$\pm$0.12 & 93.96$\pm$0.03 & 76.28$\pm$0.06 & 93.31$\pm$0.02 \\
			ResNet-101 & 77.78$\pm$0.14 & 94.39$\pm$0.04 & 76.85$\pm$0.05 & 93.93$\pm$0.02 \\
			ResNet-152 & 77.69$\pm$0.13 & 94.19$\pm$0.05 & 76.78$\pm$0.05 & 93.55$\pm$0.02 \\
			ResNeXt-50 & 82.07$\pm$0.12 & 94.96$\pm$0.04 & 81.03$\pm$0.06 & 94.52$\pm$0.03 \\
			ResNeXt101 & 82.36$\pm$0.10 & 95.01$\pm$0.03 & 81.35$\pm$0.05 & 94.58$\pm$0.02 \\
			wide\_ResNet50 & 80.75$\pm$0.11 & 94.73$\pm$0.05 & 79.77$\pm$0.06 & 94.21$\pm$0.03 \\
			wide\_ResNet101 & 81.34$\pm$0.12 & 94.85$\pm$0.04 & 80.46$\pm$0.06 & 94.33$\pm$0.02 \\
			\bottomrule
			\end{tabular}}
\end{minipage}
\end{table}

\begin{table}[!htb]
\begin{minipage}{.48\linewidth}
    \centering

    \caption{Comparison results on the \textit{ImageNet}.}
    \label{tab: exp_imagenet}

	%\normalem
	\renewcommand\tabcolsep{6.0pt}
	%\medskip
	\scalebox{0.6}{
		\begin{tabular}{l|c|c|c|c}
			\toprule
			\multicolumn{1}{l|}{\multirow{2}{*}{Backbone}} & \multicolumn{2}{c|}{Supervised} & \multicolumn{2}{c}{LBE}                     \\ \cline{2-5}  
			& top-1 & top-5 & top-1 & top-5 \\ 	\midrule
			
				ResNet-18       & 69.57$\pm$0.18 & 89.24$\pm$0.08 & 67.37$\pm$0.11 & 88.53$\pm$0.04  \\
ResNet-34       & 73.27$\pm$0.16 & 91.26$\pm$0.07 & 71.75$\pm$0.09 & 90.38$\pm$0.03 \\
ResNet-50       & 75.99$\pm$0.17 & 92.98$\pm$0.07 & 74.82$\pm$0.08 & 92.15$\pm$0.03 \\
ResNet-101      & 77.56$\pm$0.15 & 93.79$\pm$0.06 & 76.65$\pm$0.08  & 93.07$\pm$0.03  \\
ResNet-152      & 77.84$\pm$0.16 & 93.84$\pm$0.07 & 76.91$\pm$0.07  & 93.23$\pm$0.02 \\
ResNeXt-50      & 77.78$\pm$0.16 & 93.39$\pm$0.07 & 76.88$\pm$0.07 & 92.81$\pm$0.03 \\
ResNeXt101      & 78.77$\pm$0.15 & 94.28$\pm$0.06 & 77.83$\pm$0.06  & 93.91$\pm$0.03  \\
wide\_ResNet50  & 78.08$\pm$0.17 & 93.97$\pm$0.08 & 77.18$\pm$0.06 & 93.32$\pm$0.04 \\
wide\_ResNet101 & 78.29$\pm$0.16 & 94.08$\pm$0.07 & 77.21$\pm$0.06 & 93.55$\pm$0.03 \\ 
			\bottomrule
			\end{tabular}}
\end{minipage}\hfill
\begin{minipage}{.5\linewidth}
    \centering

    \caption{Results on the \textit{Tiny-ImageNet}.}
    \label{tab: exp_tiny}

    %\medskip
	\scalebox{0.6}{
		\begin{tabular}{c|c|c|c|c}
			\toprule
			\multicolumn{1}{c|}{\multirow{2}{*}{Backbone}} & \multicolumn{2}{c|}{Supervised} & \multicolumn{2}{c}{LBE}                     \\ \cline{2-5}  
			& top-1 & top-5 & top-1 & top-5 \\ 	\midrule
			ResNet-18       & 58.55$\pm$0.16 & 78.95$\pm$0.07 & 56.96$\pm$0.10 & 77.94$\pm$0.03 \\
ResNet-34       & 61.34$\pm$0.15 & 79.62$\pm$0.06 & 60.02$\pm$0.09 & 78.35$\pm$0.03 \\
ResNet-50       & 61.12$\pm$0.15 & 79.32$\pm$0.05 & 60.35$\pm$0.08 & 78.12$\pm$0.02  \\
ResNet-101      & 62.45$\pm$0.14 & 80.87$\pm$0.05 & 60.76$\pm$0.08 & 79.02$\pm$0.02 \\
ResNet-152      & 61.86$\pm$0.15 & 79.58$\pm$0.04 & 60.43$\pm$0.09 & 78.23$\pm$0.03 \\
ResNeXt-50      & 62.67$\pm$0.15 & 80.95$\pm$0.05 & 61.13$\pm$0.09 & 78.98$\pm$0.03  \\
ResNeXt101      & 63.01$\pm$0.14 & 81.13$\pm$0.05 & 61.33$\pm$0.08 & 79.35$\pm$0.02  \\
wide\_ResNet50  & 62.13$\pm$0.15 & 79.95$\pm$0.06 & 60.52$\pm$0.10 & 78.67$\pm$0.03  \\
wide\_ResNet101 & 62.34$\pm$0.15 & 80.36$\pm$0.04 & 60.85$\pm$0.09 & 78.85$\pm$0.02 \\
			\bottomrule
			\end{tabular}}
\end{minipage}
\end{table}

Table~\blue{\ref{tab: exp_c10}},~\blue{\ref{tab: exp_c100}} shows the top-1 and top-5 classification accuracy (\%) of supervised learning and LBE on CIFAR-10 dataset. 
From this table, we make the following observations. 
\textbf{First}, our method achieves competitive performance compared to the supervised learning baseline in 7 out of the 9 backbones and achieves a smaller standard deviation of averaged top-1 and top-5 accuracy than the baselines. 
This demonstrates the effectiveness of our method.
Our method learns to retrieve a set of training examples that are similar to the query examples and predicts labels for query examples by using the class labels of the retrieved ones. The Siamese network calculates similarities between the query and all training examples for retrieving similar training examples. The matching network calculates the similarity between the query and retrieved examples to predict the class label for the query. The most important similarity matrix is updated to generate ``ground truth'' similarity between training examples. Instead of using the specific similarity between examples, the supervised learning method is just using the general class label of training examples to train the network, which might cause the trained model to overfit the training set easily and harder to generalize well to new data. \textbf{Second}, our method achieves the best performance using the ResNeXt101 backbone, which further demonstrates the effectiveness of LBE in driving the frontiers of image classification forward.

Table~\blue{\ref{tab: exp_imagenet}},~\blue{\ref{tab: exp_tiny}} shows the top-1 and top-5 classification accuracy (\%) of supervised learning and LBE on ImageNet and Tiny-ImageNet datasets.
Applying our proposed LBE method to each backbone, we can achieve comparable, even better classification performance on the test set compared to the baselines. 
Especially when we use the ResNeXt101 backbone, our method achieves the best performance among all 
backbones.
This further demonstrates the effectiveness of learning by examples.

\vspace{-0.5em}
\subsubsection{Omniglot \& miniImageNet \& tieredImageNet}
\vspace{-0.5em}
We closely follow the few-shot setting same as \blue{\cite{vinyals2016matching}} and report the accuracy of 5-way 1-shot, 5-way 5-shot, 20-way 1-shot, and 20-way 5-shot experiments on the Omniglot dataset in Table \blue{\ref{tab: exp_omniglot}}.
As can be seen, we make two observations: 
First, under the same-way setting, our LBE can achieve comparable performance in the 1-shot setting and perform better results in terms of the 5-shot case. 
Second, under the same-shot setting, given more classes, the increasing gap between our LBE and MatchingNet becomes larger, which demonstrates the advantage of our LBE on retrieving a set of examples automatically from the given support set that is similar to the query examples in the tested batch set. 
% \red{TODO: add experimental analysis}
Table~\blue{\ref{tab: exp_miniImageNet3}},~\blue{\ref{tab: exp_miniImageNet1}},~\blue{\ref{tab: exp_miniImageNet2}} report the five-way classification results on miniImageNet dataset using various networks. 
In the 5 way 5-shot setting using WRN-28-10, our LBE (81.78\%) outperforms MatchingNet~\blue{\cite{vinyals2016matching}} (76.32\%) by a large margin, \textit{i.e.}, 5.46\%. 
And our LBE also can achieve comparable results with SimpleShot~\blue{\cite{wang2019simpleshot}}, although we do not need any feature normalization tricks used in SimpleShot. 
This also validates the rationality of our LBE on retrieving a set of examples from the given support set that are similar to the query examples in the tested batch set. 
% Then we use labels of the retrieved examples to predict the label for the tested batch set. 

For the few-shot setting on the tieredImageNet dataset, we also implement the same network architecture (ResNet-18, MobileNet, WRN-28-10) as SimpleShot~\blue{\cite{wang2019simpleshot}} in our Siamese and matching network.
The results are reported in Table~\blue{\ref{tab: exp_tiredImageNet}}.
Our LBE framework outperforms most baselines in both few-shot settings using those above networks. 
Note that when using WRN-28-10, our LBE can outperform the SimpleShot by 1.28\%,  in terms of the 5-way 1-shot setting. 
This infers the importance of the Siamese and matching network in our LBE framework for the few-shot learning.

\begin{table}[!htb]
    \centering

    \caption{Results on the \textit{Omniglot} dataset.}
    \label{tab: exp_omniglot}

	%\normalem
	\renewcommand\tabcolsep{6.0pt}
	%\medskip
	\scalebox{0.8}{
		\begin{tabular}{lllll}
			\toprule
			\multicolumn{1}{l}{\multirow{2}{*}{Model}} & \multicolumn{2}{c}{5 way (\%)} & \multicolumn{2}{c}{20 way (\%)}                     \\   
			& 1-shot & 5-shot & 1-shot & 5-shot \\ 	\midrule
MANN~\blue{\cite{adam2016metalearning}}                               & 82.80                             & 94.90                  &               -        &       -                \\
Siamese Net~\blue{\cite{koch2015siamese}}        & 97.30                             & 98.40                  & 88.10                  & 97.00                    \\
MAML~\blue{\cite{finn2017modelagnostic}}                     & 98.70                  & 99.90                  & 95.80                  & 98.90                  \\
VAMPIRE~\blue{\cite{nguyen2019uncertainty}}                     & 96.27                 & 98.77                  & 86.60                  & 96.14                  \\
ProtoNet~\blue{\cite{snell2017prototypical}}                     & 98.80                  & 99.70                  & 96.00                  & 98.90                  \\
RelationNet~\blue{\cite{sung2018learning}}                     & 99.60                  & 99.80                  & 97.60                  & 99.10                  \\
MAML++~\blue{\cite{antoniou2019how}}                     & 99.47                 & -                  & 97.65                  & 99.33                  \\
MatchingNet~\blue{\cite{vinyals2016matching}}                     & 98.10                    & 98.90                  & 93.80                  & 98.50                  \\
LBE (w/o weight tying)                                           & 98.93 ($\uparrow$0.83)& 99.35 ($\uparrow$0.45) & 97.07 ($\uparrow$3.27) & 99.01 ($\uparrow$0.51)         \\
LBE (w. weight tying)                          & \textbf{99.23 ($\uparrow$1.13)} & \textbf{99.61 ($\uparrow$0.71)} & \textbf{97.25 ($\uparrow$3.45)}  & \textbf{99.22 ($\uparrow$0.72)} \\
			\bottomrule
			\end{tabular}}
\end{table}

\begin{table}[!htb]
    \centering

    \caption{Results on the \textit{miniImageNet} dataset for five-way classification using various networks.}
    \label{tab: exp_miniImageNet3}

	%\normalem
	\renewcommand\tabcolsep{6.0pt}
	%\medskip
	\scalebox{0.65}{
		\begin{tabular}{lccc|lccc}
			\toprule
			Method            & Network & 1-shot & 5-shot & Method            & Network & 1-shot & 5-shot \\
			\midrule
Qiao \textit{et al.}~\blue{\cite{qiao2018few}}       & WRN-28-10                         & 59.60$\pm$0.41               & 73.74$\pm$0.19               & SimpleShot~\blue{\cite{wang2019simpleshot}} (UN)   & MobileNet                   & 55.70$\pm$0.20               & 77.46$\pm$0.15               \\
MatchingNet~\blue{\cite{vinyals2016matching}}       & WRN-28-10                         & 64.03$\pm$0.20               & 76.32$\pm$0.16               & SimpleShot~\blue{\cite{wang2019simpleshot}} (L2N)  & MobileNet                   & 59.43$\pm$0.20               & 78.00$\pm$0.15               \\
LEO~\blue{\cite{rusu2019metalearning}}               & WRN-28-10                         & 61.76$\pm$0.08               & 77.59$\pm$0.12               & SimpleShot~\blue{\cite{wang2019simpleshot}} (CL2N) & MobileNet                   & 61.30$\pm$0.20              & 78.37$\pm$0.15               \\
 ProtoNet~\blue{\cite{snell2017prototypical}}          & WRN-28-10                         & 62.60$\pm$0.20               & 79.97$\pm$0.14  & LBE (w/o weight tying)               & MobileNet                   & 62.68$\pm$0.20               & 79.73$\pm$0.14               \\
FEAT~\blue{\cite{ye2020fewshot}}              & WRN-28-10                         & 65.10$\pm$0.20               & 81.11$\pm$0.14               & LBE (w. weight tying)      & MobileNet                   & \textbf{62.81$\pm$0.19}      & \textbf{80.04$\pm$0.13}      \\ 
SimpleShot~\blue{\cite{wang2019simpleshot}} (UN)  & WRN-28-10                         & 57.26$\pm$0.21 & 78.99$\pm$0.14                &  SimpleShot~\blue{\cite{wang2019simpleshot}} (UN)   & DenseNet                    & 57.81$\pm$0.21               & 80.43$\pm$0.15               \\
SimpleShot~\blue{\cite{wang2019simpleshot}} (L2N)  & WRN-28-10                         & 61.22$\pm$0.21 & 81.00$\pm$0.14               &  SimpleShot~\blue{\cite{wang2019simpleshot}} (L2N) & DenseNet                    & 61.49$\pm$0.20               & 81.48$\pm$0.14               \\
SimpleShot~\blue{\cite{wang2019simpleshot}} (CL2N)  & WRN-28-10                         & 63.50$\pm$0.20               & 80.33$\pm$0.14                &  SimpleShot~\blue{\cite{wang2019simpleshot}} (CL2N)  & DenseNet                    & 64.29$\pm$0.20               & 81.50$\pm$0.14               \\
Dhillon \textit{et al.}~\blue{\cite{Dhillon2020a}} & WRN-28-10                         & 65.73$\pm$0.68               &  78.40$\pm$0.52               & LBE (w/o weight tying)               & DenseNet                    & 66.36$\pm$0.18               & 82.74$\pm$0.12               \\
SIB~\blue{\cite{Hu2020Empirical}} & WRN-28-10                         & 70.00$\pm$0.60               & 79.20$\pm$0.40               &  LBE (w. weight tying)      & DenseNet                    & \textbf{66.82$\pm$0.17}      & \textbf{83.91$\pm$0.11}    \\
BD-CSPN~\blue{\cite{liu2020prototype}} & WRN-28-10                         & 70.31$\pm$0.93               & 81.89$\pm$0.60               &  LBE (w/o weight tying)                & WRN-28-10                         & 72.92$\pm$0.18               & 83.45$\pm$0.12               \\
EPNet~\blue{\cite{rodriguez2020embedding}} & WRN-28-10                         & 70.74$\pm$0.85               & 84.34$\pm$0.53               &  LBE (w. weight tying)      & WRN-28-10                         & \textbf{73.23$\pm$0.17}      & \textbf{83.78$\pm$0.12}  \\
			\bottomrule
			\end{tabular}}
\end{table}

\begin{table}[!htb]
\begin{minipage}{.48\linewidth}
    \centering

    \caption{Results on the \textit{miniImageNet} dataset for five-way classification using Conv-4 network.}
    \label{tab: exp_miniImageNet1}

	%\normalem
	\renewcommand\tabcolsep{6.0pt}
	%\medskip
	\scalebox{0.65}{
		\begin{tabular}{lccc}
			\toprule
			Model & Network & 1-shot & 5-shot                  \\ \midrule
			Meta LSTM~\blue{\cite{ravi2017optimization}}         & Conv-4 & 43.44$\pm$0.77        & 60.60$\pm$0.71                              \\
MatchingNet~\blue{\cite{vinyals2016matching}}       & Conv-4 & 43.56$\pm$0.84                              & 55.31$\pm$0.73                              \\
MAML~\blue{\cite{finn2017modelagnostic}}              & Conv-4 & 48.70$\pm$1.84                              & 63.11$\pm$0.92          \\
ProtoNet~\blue{\cite{snell2017prototypical}}          & Conv-4 & 49.42$\pm$0.78                              & 68.20$\pm$0.66          \\
Reptile~\blue{\cite{nichol2018first}}           & Conv-4 & 49.97$\pm$0.32                              & 65.99$\pm$0.58         \\
mAP-SSVM~\blue{\cite{triantafillou2017fewshot}}          & Conv-4 & 50.32$\pm$0.80                              & 63.94$\pm$0.72                              \\
GNN~\blue{\cite{garcia2018fewshot}}               & Conv-4 & 50.33$\pm$0.36                              & 66.41$\pm$0.63                              \\
RelationNet~\blue{\cite{sung2018learning}}       & Conv-4 & 50.44$\pm$0.82                              & 65.32$\pm$0.70                              \\
Meta SGD~\blue{\cite{li2017metasgd}}          & Conv-4 & 50.47$\pm$1.87                              & 64.03$\pm$0.94                              \\
Qiao \textit{et al.}~\blue{\cite{qiao2018few}}       & Conv-4 & 54.53$\pm$0.40                              & 67.87$\pm$0.20                              \\
TPN~\blue{\cite{liu2019learning}}  & Conv-4 & 53.75$\pm$0.86                              & 69.43$\pm$0.67                              \\
CC+rot~\blue{\cite{gidaris2019boosting}}   & Conv-4 & 54.83$\pm$0.43                             & 71.86$\pm$0.33                             \\
SimpleShot~\blue{\cite{wang2019simpleshot}} & Conv-4 & 49.69$\pm$0.19                              & 66.92$\pm$0.17                              \\
LBE (w/o weight tying)              & Conv-4 & 54.65$\pm$0.15          & 71.23$\pm$0.14          \\
LBE (w. weight tying)      & Conv-4 & \textbf{55.13$\pm$0.13} & \textbf{71.92$\pm$0.12} \\
			\bottomrule
			\end{tabular}}
% 			\vspace{-0.5em}
\end{minipage}\hfill
\begin{minipage}{.5\linewidth}
    \centering

    \caption{Results on the \textit{miniImageNet} dataset for five-way classification using ResNet-18/15/12.}
    \label{tab: exp_miniImageNet2}

    %\medskip
	\scalebox{0.65}{
		\begin{tabular}{lccc}
			\toprule
			Model & Network & 1-shot & 5-shot                  \\ \midrule
			MAML~\blue{\cite{finn2017modelagnostic}}              & ResNet-18 & 49.61$\pm$0.92          & 65.72$\pm$0.77                              \\
Chen \textit{et al.}~\blue{\cite{chen2019a}}       & ResNet-18 & 51.87$\pm$0.77          & 75.68$\pm$0.63                              \\
RelationNet~\blue{\cite{sung2018learning}}       & ResNet-18 & 52.48$\pm$0.86          & 69.83$\pm$0.68          \\
MatchingNet~\blue{\cite{vinyals2016matching}}       & ResNet-18 & 52.91$\pm$0.88          & 68.88$\pm$0.69          \\
ProtoNet~\blue{\cite{snell2017prototypical}}          & ResNet-18 & 54.16$\pm$0.82          & 73.68$\pm$0.65          \\
Gidaris \textit{et al.}~\blue{\cite{gidaris2018dynamic}}    & ResNet-15 & 55.45$\pm$0.89          & 70.13$\pm$0.68                              \\
SNAIL~\blue{\cite{mishra2018a}}             & ResNet-15 & 55.71$\pm$0.99          & 68.88$\pm$0.92                              \\
Bauer \textit{et al.}~\blue{\cite{bauer2017discriminative}}      & ResNet-34 & 56.30$\pm$0.40          & 73.90$\pm$0.30                              \\
adaCNN~\blue{\cite{munkhdalai2018rapid}}            & ResNet-15 & 56.88$\pm$0.62          & 71.94$\pm$0.57                              \\
TADAM~\blue{\cite{oreshkin2018tadam}}             & ResNet-15 & 58.50$\pm$0.30          & 76.70$\pm$0.30                              \\
CAML~\blue{\cite{jiang2019learning}}              & ResNet-12 & 59.23$\pm$0.99          & 72.35$\pm$0.71                              \\
Dhillon \textit{et al.}~\blue{\cite{Dhillon2020a}}  & ResNet-12 & 62.35$\pm$0.66         & 74.53$\pm$0.54                             \\
SimpleShot~\blue{\cite{wang2019simpleshot}} & ResNet-18 & 62.85$\pm$0.20          & 80.02$\pm$0.14         \\
LBE(w/o weight tying)                & ResNet-18 & 64.26$\pm$0.18          & 81.23$\pm$0.12          \\
LBE (w. weight tying)      & ResNet-18 & \textbf{64.48$\pm$0.16} & \textbf{81.41$\pm$0.11} \\
			\bottomrule
			\end{tabular}}
% 			\vspace{-0.5em}
\end{minipage}
\end{table}

\begin{table}[!htb]
    \centering

    \caption{Results on the \textit{tiredImageNet} dataset for five-way classification using various networks.}
    \label{tab: exp_tiredImageNet}

	%\normalem
	\renewcommand\tabcolsep{6.0pt}
	%\medskip
	\scalebox{0.65}{
		\begin{tabular}{lccc|lccc}
			\toprule
			Method            & Network & 1-shot & 5-shot & Method            & Network & 1-shot & 5-shot \\
			\midrule
MAML~\blue{\cite{finn2017modelagnostic}}    & Conv-4                      & 51.67$\pm$1.81               & 70.30$\pm$0.08              & MetaOptNet~\blue{\cite{lee2019meta}}  & ResNet-18                   & 65.99$\pm$0.72               & 81.56$\pm$0.53              \\
  Reptile~\blue{\cite{nichol2018first}}           & Conv-4                      & 48.97$\pm$0.21               & 66.47$\pm$0.21               &
  SimpleShot~\blue{\cite{wang2019simpleshot}}  & ResNet-18                   & 69.09$\pm$0.22               & 84.58$\pm$0.16               \\
SimpleShot~\blue{\cite{wang2019simpleshot}} (CL2N) & Conv-4                      & 51.02$\pm$0.20               & 68.98$\pm$0.18               &
LBE (w/o weight tying)               & ResNet-18                   & 70.11$\pm$0.21               & 85.38$\pm$0.15               \\
LBE (w/o weight tying)               & Conv-4                      & 51.85$\pm$0.18               & 70.39$\pm$0.16               & LBE (w. weight tying)      & ResNet-18                   & \textbf{70.43$\pm$0.20}      & \textbf{85.76$\pm$0.14}      \\
LBE (w. weight tying)      & Conv-4                      & \textbf{52.43$\pm$0.16}      & \textbf{71.06$\pm$0.15}      & SimpleShot~\blue{\cite{wang2019simpleshot}} (UN)   & MobileNet                   & 63.65$\pm$0.22               & 84.01$\pm$0.16               \\
Meta SGD~\blue{\cite{li2017metasgd}}          & WRN-28-10                         & 62.95$\pm$0.03               & 79.34$\pm$0.06               & SimpleShot~\blue{\cite{wang2019simpleshot}} (L2N)  & MobileNet                   & 68.66$\pm$0.23               & 85.43$\pm$0.15               \\
FEAT~\blue{\cite{ye2020fewshot}}               & WRN-28-10                         & 70.41$\pm$0.23               & 84.38$\pm$0.16               & SimpleShot~\blue{\cite{wang2019simpleshot}} (CL2N) & MobileNet                   & 69.47$\pm$0.22               & 85.17$\pm$0.15               \\
LEO~\blue{\cite{rusu2019metalearning}}               & WRN-28-10                         & 66.33$\pm$0.05               & 81.44$\pm$0.09               & LBE (w/o weight tying)               & MobileNet                   & 70.21$\pm$0.21               & 85.65$\pm$0.15               \\
CC+rot~\blue{\cite{gidaris2019boosting}}  & WRN-28-10                         & 70.53$\pm$0.51               &  84.98$\pm$0.36               & LBE (w. weight tying)      & MobileNet                   & \textbf{70.62$\pm$0.19}      & \textbf{86.02$\pm$0.14}      \\
SimpleShot~\blue{\cite{wang2019simpleshot}} (UN) & WRN-28-10 & 63.85$\pm$0.21 & 84.17$\pm$0.15 &
SimpleShot~\blue{\cite{wang2019simpleshot}} (UN)   & DenseNet                    & 64.35$\pm$0.23               & 85.69$\pm$0.15               \\
SimpleShot~\blue{\cite{wang2019simpleshot}} (L2N) & WRN-28-10 & 66.86$\pm$0.21 & 85.50$\pm$0.14 &
SimpleShot~\blue{\cite{wang2019simpleshot}} (L2N)  & DenseNet                    & 69.91$\pm$0.22               & 86.42$\pm$0.15               \\
SimpleShot~\blue{\cite{wang2019simpleshot}} (CL2N) & WRN-28-10                         & 69.75$\pm$0.20               & 85.31$\pm$0.15               &
SimpleShot~\blue{\cite{wang2019simpleshot}} (CL2N) & DenseNet                    & 71.32$\pm$0.22               & 86.66$\pm$0.15               \\
LBE (w/o weight tying)               & WRN-28-10                         & 72.93$\pm$0.19               & 85.96$\pm$0.14               &  LBE (w/o weight tying)  & DenseNet                    & 72.08$\pm$0.21              & 87.15$\pm$0.14               \\
LBE (w. weight tying)      & WRN-28-10                         & \textbf{73.12$\pm$0.18}      & \textbf{86.17$\pm$0.13}      & LBE (w. weight tying)  & DenseNet                    & \textbf{72.52$\pm$0.20}               & \textbf{87.31$\pm$0.13}               \\
			\bottomrule
			\end{tabular}}
\end{table}

% \vspace{-1.0em}
\subsubsection{Visualizations}
\begin{figure}[!htb]
% \vspace{-1em}
% \setlength{\abovecaptionskip}{-0em}
% \setlength{\belowcaptionskip}{-1em}
	    \centering
	%\begin{center}
		%\fbox{\rule{0pt}{2in} \rule{0.6\linewidth}{0pt}}
		\includegraphics[width=0.85\linewidth]{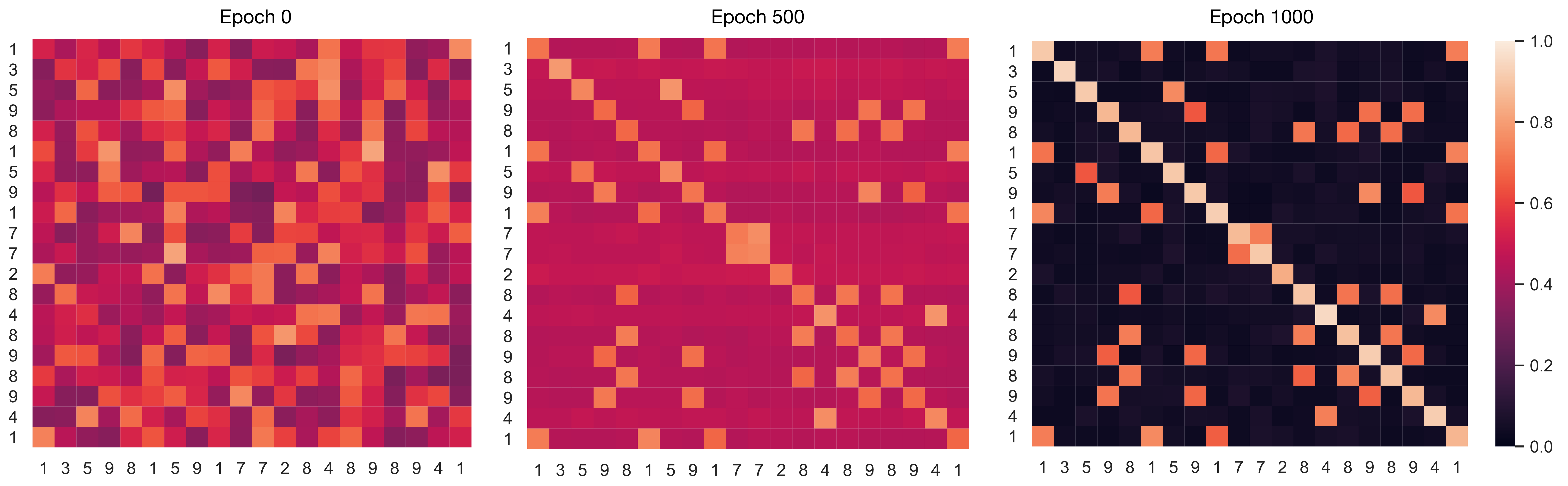}
	%\end{center}
    	\caption{Visualization of the ``groundtruth'' similarity matrix between training examples by using LBE through the training process, where we select $20$ random training examples with their class label (noted on x, y axis). The similarity value $a \in [0, 1]$, and larger value indicates more similarity between two examples.  }
	\label{fig: sim_mat}
\end{figure}

\begin{figure}[!htb]
% \vspace{-0.5em}
% \setlength{\abovecaptionskip}{-0em}
% \setlength{\belowcaptionskip}{-2em}
\centering
% \fbox{\rule{0pt}{1.8in} \rule{0.8\linewidth}{0pt}}
  \includegraphics[width=0.85\linewidth]{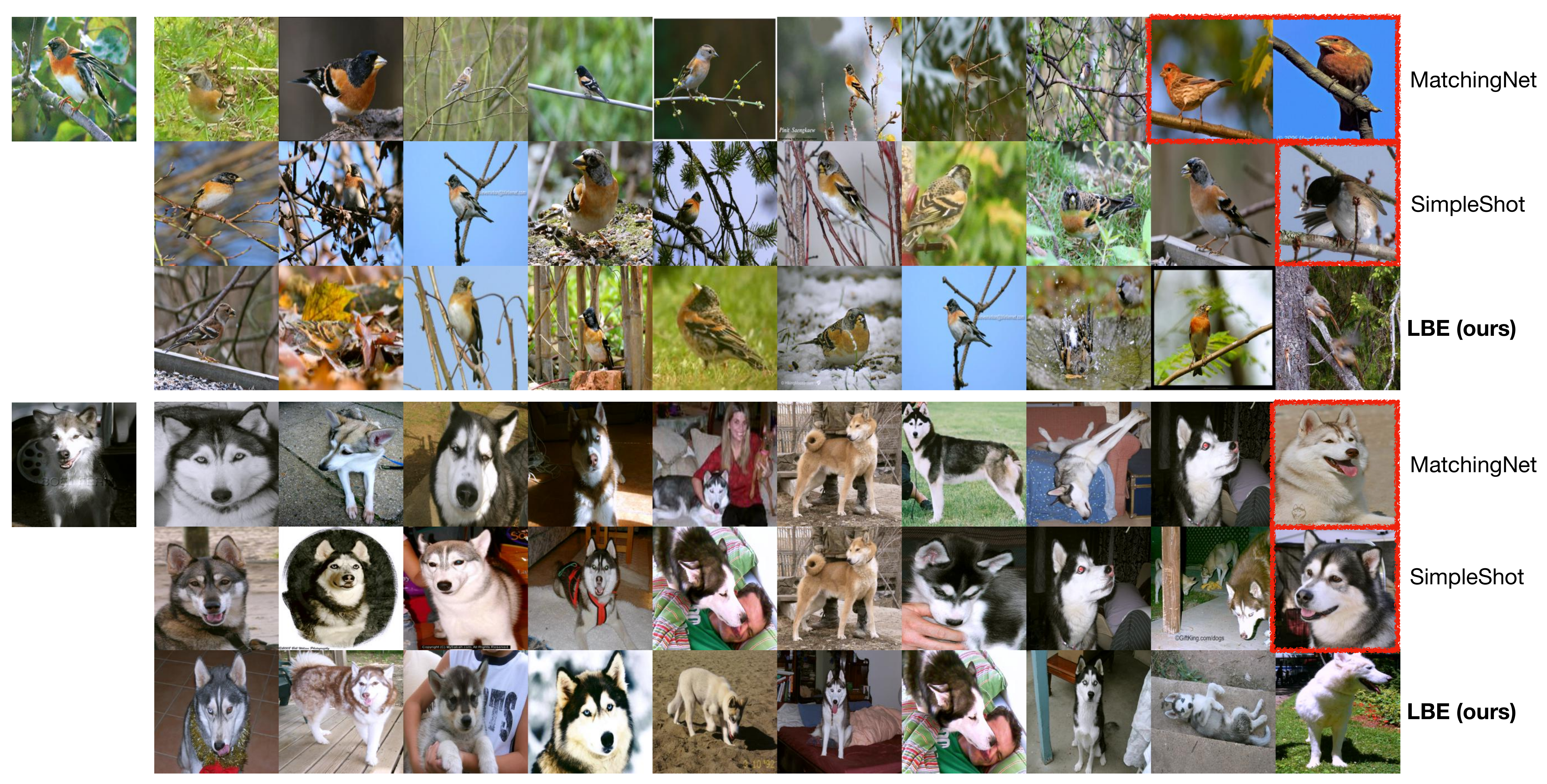}
   \caption{Top-10 retrieved images from MatchingNet, SimpleShot, and LBE (ours). Red denotes false positives.}
\label{fig: exp_visual}
\end{figure}

% \vspace{-0.5em}
To better analyze the ``ground-truth'' similarity matrix learned in our proposed LBE framework, we select $20$ random training examples and visualize their similarity matrix at three epochs (0, 500, 1000) during the training process. As can be seen in Figure \blue{\ref{fig: sim_mat}}, our proposed LBE framework captures the ``ground-truth'' similarity between training examples, which is consistent with the class label of examples. A larger similarity between the two examples means that the bigger probability that they are from the same class. The similarity between examples from different classes is smaller than $0.1$, which means that our LBE framework is capable of retrieving a set of training examples that are from the same class. This will help our proposed LBE framework automatically retrieve a set of training examples that are similar to the query examples and predict labels for query examples by using the class labels of the retrieved examples.
Given a query image, we report the top 10 retrieved images from MatchingNet~\cite{vinyals2016matching}, SimpleShot~\cite{wang2019simpleshot}, and our LBE in Figure~\blue{\ref{fig: exp_visual}}. There exist no false positives retrieved from our LBE, which further demonstrates the effectiveness of our LBE framework.

% \vspace{-0.5em}
\subsection{Ablation Study}
% \vspace{-0.5em}
In this section, we perform extensive ablation studies on fast retrieval \& matching, weight tying of $T$ and $S$, and the temperature parameter in $T$.

\begin{table}[!htb]
% \setlength{\abovecaptionskip}{-0em}
% \setlength{\belowcaptionskip}{-0.5em}
	%\normalem
	\caption{Ablation study on hash codes for the Siamese $T$ and matching $S$ network on CIFAR-10 and ImageNet dataset.}
	\label{tab: ab_fast}
	\renewcommand\tabcolsep{6.0pt}
	\centering
	\scalebox{0.75}{
		\begin{tabular}{lllllllll}
			\toprule
			 \multicolumn{1}{l}{\multirow{2}{*}{Method}} & \multicolumn{1}{l}{\multirow{2}{*}{$T$}} & \multicolumn{1}{l}{\multirow{2}{*}{$S$}} &  	  \multicolumn{6}{c}{CIFAR-10 / ImageNet}                     \\ 
			& & & top-1$\uparrow$ & top-5 $\uparrow$& train (s/b) $\downarrow$ & infer (s/b)$\downarrow$ & T (s/e)$\downarrow$ & S (s/e)$\downarrow$  \\
			
			\midrule
			LBE & & &  \textbf{92.03} (-) & \textbf{98.83} (-)	& 2.62 (-)& 1.10 (-) & 0.39 (-) & 0.28 (-) \\
			LBE-fast-$T$ & \checkmark & & 91.97 ($\downarrow$0.06)& 98.81 ($\downarrow$0.02)	 & 1.83 ($\downarrow$0.79) & 0.87 ($\downarrow$0.23) & \textbf{0.16 ($\downarrow$0.23)} & 0.28 (-)  \\
				LBE-fast-$S$ &  &\checkmark & 91.99 ($\downarrow$0.04) & 98.82 ($\downarrow$0.01) & 2.31 ($\downarrow$0.31)	& 0.96 ($\downarrow$0.14) & 0.39 (-) & \textbf{0.12 ($\downarrow$0.16)}  \\
					LBE-fast-$TS$ & \checkmark &\checkmark &  91.91 ($\downarrow$0.12) & 98.79 ($\downarrow$0.04) & \textbf{1.53 ($\downarrow$1.09)} & \textbf{0.71 ($\downarrow$0.39)} & \textbf{0.16 ($\downarrow$0.23)} & \textbf{0.12 ($\downarrow$0.16)} \\ \hline
			LBE & & &  \textbf{67.37} (-) & \textbf{88.53} (-) & 32.27 (-) & 16.65 (-) & 9.86 (-) & 6.83 (-)  \\
			LBE-fast-$T$ & \checkmark & & 67.19 ($\downarrow$0.18)& 88.47 ($\downarrow$0.06)& 15.95 ($\downarrow$16.32) & 10.26 ($\downarrow$6.39) & \textbf{1.98 ($\downarrow$7.88)} & 6.83 (-)  \\
				LBE-fast-$S$ &  &\checkmark & 67.25 ($\downarrow$0.12)	& 88.49 ($\downarrow$0.04)	& 17.12 ($\downarrow$15.15) & 12.35 ($\downarrow$4.30) & 9.86 (-) & \textbf{1.75 ($\downarrow$5.08)}  \\
					LBE-fast-$TS$ & \checkmark &\checkmark &  67.12 ($\downarrow$0.25)& 88.43 ($\downarrow$0.10)& \textbf{11.13 ($\downarrow$21.14)} & \textbf{7.53 ($\downarrow$9.12)}	& \textbf{1.98 ($\downarrow$7.88)} & \textbf{1.75 ($\downarrow$5.08)}  \\
			\bottomrule
			\end{tabular}}
\end{table}

\textbf{Fast retrieval \& matching.}
Table~\blue{\ref{tab: ab_fast}} reports the results of the ablation study on applying hash codes to our LBE framework on the CIFAR-10 and ImageNet datasets.
For training and inference speed per batch, our LBE-fast-$T$ model can achieve 50.57\% and 38.38\% faster than the vanilla LBE, validating the effectiveness of hash codes. 
Similar to LBE-fast-$T$, our LBE-fast-$S$ and LBE-fast-$TS$ model achieve comparable performance to the vanilla LBE in terms of top-1 and top-5 accuracy, which shows the robustness of applying hash codes to both the matching and Siamese networks. 
In terms of the time of calculating the similarity between a test example and all training examples, applying hash codes to both networks increase the speed of Siamese network $T$ and matching network $S$ by 79.91\% and 74.37\%. 
This further demonstrates the superiority of hash codes in fast retrieval and matching.

\begin{table}[!htb]
\begin{minipage}{.6\linewidth}
    \centering

    \caption{Ablation study on the weight tying.} 
    %between the Siamese and matching network.}
    \label{tab: ab_share}

    %\medskip
	\scalebox{0.7}{
		\begin{tabular}{ccccc}
			\toprule
			Siamese & matching & weight tying? & top-1 & top-5 \\ 	
			\midrule
			ResNet-18 & ResNet-18 &            & 91.92$\pm$0.09 & 98.78$\pm$0.04 \\  
			ResNet-18 & ResNet-18 & \checkmark & 92.03$\pm$0.06 & 98.83$\pm$0.03 \\ 
			ResNet-34 & ResNet-34 &            & 92.05$\pm$0.08 & 98.91$\pm$0.05 \\ 
			ResNet-34 & ResNet-34 & \checkmark & 92.31$\pm$0.06 & 98.96$\pm$0.03 \\ 
			ResNet-50 & ResNet-50 &            &   92.23$\pm$0.07 & 99.07$\pm$0.04 \\  
			ResNet-50 & ResNet-50 & \checkmark &   92.52$\pm$0.05 & 99.11$\pm$0.03 \\ 
			ResNet-101 & ResNet-101 &           &  92.56$\pm$0.08 & 99.16$\pm$0.03 \\  
			ResNet-101 & ResNet-101 & \checkmark & \textbf{92.83}$\pm$\textbf{0.05} & \textbf{99.22}$\pm$\textbf{0.02} \\ 
			\midrule
			ResNet-18 & ResNet-34 &   & 92.08$\pm$0.08 & 98.84$\pm$0.05 \\ 
			ResNet-18 & ResNet-50 &   & 92.17$\pm$0.10 & 98.88$\pm$0.04 \\ 
			ResNet-18 & ResNet-101 &  & 92.29$\pm$0.09 & 98.93$\pm$0.05 \\ 
			ResNet-34 & ResNet-18 &   & 92.13$\pm$0.09 & 98.87$\pm$0.04 \\ 
			ResNet-50 & ResNet-18 &   & 92.21$\pm$0.08 & 98.90$\pm$0.04 \\ 
			ResNet-101 & ResNet-18 &  & \textbf{92.32}$\pm$ \textbf{0.07} & \textbf{98.95}$\pm$\textbf{0.03} \\
			\bottomrule
			\end{tabular}}

\end{minipage}\hfill
\begin{minipage}{.4\linewidth}
    \centering

    \caption{Ablation study on the temperature.}
    \label{tab: ab_temp}

	%\normalem
	%\caption{Exploration study on the temperature parameter in the Siamese network.}
	%\label{tab: ab_temp}
	\renewcommand\tabcolsep{6.0pt}
	%\centering
	%\begin{center}
	%\medskip
	\scalebox{0.7}{
		\begin{tabular}{ccc}
			\toprule
			temperature & top-1 & top-5 \\ 	
			\midrule
			10000 & 90.03$\pm$0.11 & 98.31$\pm$0.06 \\ 
			1000 &  90.45$\pm$0.11 & 98.42$\pm$0.06 \\ 
			100 &  91.13$\pm$0.11 & 98.54$\pm$0.05 \\ 
			10 & 91.32$\pm$0.10 & 98.61$\pm$0.05 \\ 
			1 &  91.52$\pm$0.09 & 98.68$\pm$0.05 \\ 
			0.1 & 91.67$\pm$0.08 & 98.72$\pm$0.04 \\
			0.01 &  91.84$\pm$0.08 & 98.76$\pm$0.04 \\ 
			0.001 & 91.89$\pm$0.07 & 98.77$\pm$0.04 \\ 
			0.0001 &  \textbf{92.03}$\pm$\textbf{0.06} & \textbf{98.83}$\pm$\textbf{0.03} \\
			0.00001 & 91.98$\pm$\textbf{0.06} & 98.81$\pm$\textbf{0.03} \\ 
			\bottomrule

			\end{tabular}}
\end{minipage}
\end{table}

% \vspace{-0.5em}
\textbf{Weight tying.} Table \blue{\ref{tab: ab_share}} (Top rows) explores the effect of weight tying between the Siamese network and matching network on the final performance score. As can be seen, applying weight tying to each backbone can boost the top-1 classification accuracy further. To analyze the comprehensive contribution of the Siamese and matching network, we also apply different backbones to the Siamese and matching network, as shown in Table \blue{\ref{tab: ab_share}} (Bottom rows). We can observe that the Siamese network with a better backbone achieves larger improvements over the matching network. This further validates the effectiveness of the Siamese network in training to match the ``ground-truth'' similarity between training examples. 
%\red{TO DO}

\textbf{Temperature.} Using the ResNet-18 backbone, we also explore the effect of the temperature parameter in the Siamese network on the final accuracy in Table \blue{\ref{tab: ab_temp}}. A larger temperature means less contribution to the overall loss. We can see that larger temperature, the top-1 accuracy will decrease a lot. This further demonstrates the rationality of the Siamese network in our proposed LBE framework. 
% \vspace{-0em}

\begin{table}[!htb]
\begin{minipage}{.5\linewidth}
    \centering

    \caption{Exploration study on \# of examples.}
    \label{tab: ab_example}

    %\medskip
	\scalebox{0.6}{
		\begin{tabular}{cccccc}
			\toprule
			\multicolumn{1}{c}{\multirow{2}{*}{train}} & \multicolumn{1}{c}{\multirow{2}{*}{val}}& \multicolumn{2}{c}{Supervised} & \multicolumn{2}{c}{LBE}                     \\ 
			& & top-1 & top-5 & top-1 & top-5 \\ 
% 			train & val & top-1 & top-5 \\

			\midrule
			500 &  50 &   83.65$\pm$0.11 & 97.13$\pm$0.08 & 82.52$\pm$0.09 & 96.82$\pm$0.04 \\ 
			800 &  80 &   88.25$\pm$0.09 & 97.74$\pm$0.07 & 87.68$\pm$0.08 & 97.14$\pm$0.04 \\ 
			1000 &  100 & 90.82$\pm$0.10 & 98.52$\pm$0.06 & 89.83$\pm$0.07 & 97.85$\pm$\textbf{0.03} \\ 
			2000 &  200 & 91.92$\pm$\textbf{0.08} & 99.02$\pm$\textbf{0.05} & 91.05$\pm$0.06 & 98.05$\pm$\textbf{0.03} \\ 
			3000 &  300 & 91.95$\pm$\textbf{0.08} & 99.04$\pm$0.06 & 91.65$\pm$\textbf{0.05} & 98.43$\pm$0.04 \\ 
			4000 &  400 & 92.51$\pm$0.09 & 99.16$\pm$0.07 & 91.18$\pm$0.07 & 98.56$\pm$0.05 \\ 
			4500 &  500 &  \textbf{92.95}$\pm$0.11 & \textbf{99.25}$\pm$0.08 & \textbf{92.03}$\pm$0.06 & \textbf{98.83}$\pm$\textbf{0.03} \\ 
			\bottomrule
			\end{tabular}}
\end{minipage}\hfill
\begin{minipage}{.5\linewidth}
    \centering

    \caption{Exploration study on the batch size.}
    \label{tab: ab_batch}

    %\medskip
    \scalebox{0.6}{
		\begin{tabular}{ccccc}
			\toprule
			\multicolumn{1}{c}{\multirow{2}{*}{batch size}} & \multicolumn{2}{c}{Supervised} & \multicolumn{2}{c}{LBE}                     \\
			& top-1 & top-5 & top-1 & top-5 \\	
			\midrule
			16 &  89.56$\pm$0.11 & 98.13$\pm$0.08 & 89.21$\pm$0.09 & 98.46$\pm$0.04 \\ 
			24 &  90.91$\pm$0.09 & 98.67$\pm$0.08 & 90.47$\pm$0.07 & 98.53$\pm$0.03 \\ 
			32 &  91.25$\pm$0.08 & 98.75$\pm$0.07 & 91.05$\pm$0.05 & 98.62$\pm$0.04 \\ 
			64 &  92.34$\pm$\textbf{0.05} & 99.06$\pm$0.05 & 91.73$\pm$\textbf{0.04} & 98.75$\pm$0.03 \\ 
			128 & 92.65$\pm$0.06 & 99.18$\pm$\textbf{0.04} & 92.01$\pm$0.05 & 98.82$\pm$\textbf{0.02} \\ 
			256 & \textbf{92.95}$\pm$0.11 & \textbf{99.25}$\pm$0.08 & \textbf{92.03}$\pm$0.06 & \textbf{98.83}$\pm$0.03 \\ 
			512 & 92.83$\pm$0.09 & 99.21$\pm$\textbf{0.04} & 91.96$\pm$0.06 & 98.81$\pm$0.04 \\ 
			1024 &92.46$\pm$0.06 & 99.12$\pm$0.05 & 91.84$\pm$0.05 & 98.76$\pm$0.03 \\ 
			\bottomrule
			\end{tabular}}
\end{minipage}
\end{table}

\textbf{Number of examples.} Table \blue{\ref{tab: ab_example}} explores the effect of the number of training and validation examples on the final score of LBE. As can be seen, retrieving a set of training examples that are similar to the query examples in a bigger training and validation examples pool further help the matching network predict the class label for the query examples by using the ``ground-truth'' label from retrieved examples. Our LBE framework achieves competitive top-1 and top-5 accuracy compared to the supervised learning methods.

\textbf{Batch size.} As shown in Table \blue{\ref{tab: ab_batch}}, we also ablate the effect of the batch size on the performance of our proposed LBE framework and the supervised learning method. We can observe that the top-1 and top-5 classification accuracy of our LBE framework decreases with the increase of batch size by a smaller margin compared to the supervised approach. This validates the robustness of our LBE framework to the choice of batch size, which will save us memories given limited GPU resources.

% \vspace{-1.5em}
\section{Related Work}\label{related_work}
% \vspace{-0.5em}

\textbf{Learning similarities.}
Learning similarities plays a crucial role in the supervised machine learning literature. 
The aim of similarity learning is to learn a siamese network that measures how similar or associative two objects are. 
Generally, there are four common types of similarity and metric distance learning, including classification~\blue{\cite{vinyals2016matching,koch2015siamese,andrzej2012learning}}, regression~\blue{\cite{kar2012supervised,qian2015similarity}}, ranking~\blue{\cite{gal2010large,wang2014learning}} and locality sensitive hashing~\blue{\cite{mayur2004local,omid2021locality}}.
In this work, we focus on classification similarity learning, \textit{i.e.}, learning a similarity function to classify other objects given the class of one object.

\textbf{Leveraging data similarity for classification.}
Previous works~\blue{\cite{vinyals2016matching,koch2015siamese,melekhov2016siamese}} in metric learning achieves remarkable progress in few-shot-classification, where the model is trained to learn a mapping function that maps examples belonging to the same class close together while pulling separate classes apart. 
Similarities between features of samples from the support set and query set are computed to perform classification.
Several approaches have been proposed for learning by examples. Common supervised learning approaches (AlexNet \blue{\cite{alex2012alexnet}}, ResNet \blue{\cite{he2016resnet}}, etc) use a bi-level optimization framework to update the network weights by matching the ``ground-truth'' class label of training examples directly. 

Similar to learning similarity between examples in our LBE framework, metric learning (\blue{\cite{guillaumin2009is, brian2010metric,kostinger2012large,fatih2019deep}}) is an approach based directly on a distance metric that aims to establish similarity or dissimilarity between objects. The main purpose of metric learning is to learn a new metric to reduce the distances between samples of the same class and increase the distances between the samples of different classes. The methods in (\blue{\cite{vinyals2016matching,koch2015siamese, melekhov2016siamese}}) perform similarity learning in the training and validation dataset separately. As a result, the validation performance of the model cannot be used to guide similarity learning. These works only focus on learning by examples using a bi-level optimization framework while our work aims to retrieve a set of training examples that are similar to the query examples using a three-level optimization framework.

\paragraph{Bi-level optimization.} Our framework is based on bi-level optimization (BLO)~\cite{dempe2002foundations}. BLO has been broadly applied for neural architecture search~\cite{liu2018darts}, data reweighting~\cite{shu2019meta}, hyperparameter tuning~\cite{feurer2015initializing}, meta learning~\cite{finn2017model}, learning rate tuning~\cite{BaydinOnline17}, label correction~\cite{zhengmeta19}, etc. BLO involves two levels of optimization problems where the lower level learns model weights while the upper level learns meta parameters. BLO has been extended to multi-level optimization (MLO) which involves more than two levels of optimization problems. MLO has been applied for data generation~\cite{abs-1912-07768}, interleaving multi-task learning~\cite{ban2021interleaving}, data reweighting in domain adaptation~\cite{zhao2021learning}, explainable learning~\cite{hosseini2021learning}, human-inspired learning~\cite{xie2021skillearn}, curriculum evaluation~\cite{du2021learning}, mutual knowledge distillation~\cite{du2021smallgroup}, end-to-end knowledge distillation~\cite{sheth2021learning}, etc. 
% \vspace{-1em}
\section{Conclusions}\label{conclusion}
% \vspace{-0.5em}
In this paper, we propose a novel machine learning approach – learning by examples (LBE),
inspired by the examples-driven learning technique of humans. Our LBE framework automatically retrieves a set of training examples that are similar to the query examples and predicts labels for query examples by using the class labels of the retrieved ones. We propose a multi-level optimization framework
to formalize LBE which involves three learning stages: a Siamese network is trained to retrieve similar examples; a matching network is learned to make predictions on query examples by leveraging class labels of retrieved examples; the ``ground-truth'' similarities are updated by
minimizing the validation loss calculated using the trained Siamese and matching network. 
Experiments on various
benchmarks demonstrate the effectiveness of our method on both supervised and few-shot learning.

% \section*{References}

\bibliographystyle{plain}
\bibliography{manuscript}

\begin{thebibliography}{10}

\bibitem{antoniou2019how}
Antreas Antoniou, Harrison Edwards, and Amos Storkey.
\newblock How to train your {MAML}.
\newblock In {\em Proceedings of International Conference on Learning
  Representations}, 2019.

\bibitem{ban2021interleaving}
Hao Ban and Pengtao Xie.
\newblock Interleaving learning, with application to neural architecture
  search, 2021.

\bibitem{bauer2017discriminative}
Matthias Bauer, Mateo Rojas-Carulla, Jakub~Bartłomiej \'{S}wiątkowski,
  Bernhard Sch\"{o}lkopf, and Richard~E. Turner.
\newblock Discriminative k-shot learning using probabilistic models.
\newblock {\em arXiv preprint arXiv:1706.00326}, 2017.

\bibitem{BaydinOnline17}
Atilim~Gunes Baydin, Robert Cornish, David Mart{\'{\i}}nez{-}Rubio, Mark
  Schmidt, and Frank~D. Wood.
\newblock Online learning rate adaptation with hypergradient descent.
\newblock {\em CoRR}, abs/1703.04782, 2017.

\bibitem{fatih2019deep}
Fatih Cakir, Kun He, Xide Xia, Brian Kulis, and Stan Sclarof.
\newblock Deep metric learning to rank.
\newblock In {\em Proceedings of IEEE/CVF Conference on Computer Vision and
  Pattern Recognition}, pages 1861--1870, 2019.

\bibitem{gal2010large}
Gal Chechik, Varun Sharma, Uri Shalit, and Samy Bengio.
\newblock Large scale online learning of image similarity through ranking.
\newblock {\em Journal of Machine Learning Research}, 11(36):1109--1135, 2010.

\bibitem{chen2019a}
Wei-Yu Chen, Yen-Cheng Liu, Zsolt Kira, Yu-Chiang~Frank Wang, and Jia-Bin
  Huang.
\newblock A closer look at few-shot classification.
\newblock In {\em Proceedings of International Conference on Learning
  Representations}, 2019.

\bibitem{mayur2004local}
Mayur Datar, Nicole Immorlica, Piotr Indyk, and Vahab~S Mirrokni.
\newblock Locality-sensitive hashing scheme based on p-stable distributions.
\newblock In {\em Proceedings of the twentieth annual symposium on
  Computational geometry}, pages 253--262, 2004.

\bibitem{dempe2002foundations}
Stephan Dempe.
\newblock {\em Foundations of bilevel programming}.
\newblock Springer Science \& Business Media, 2002.

\bibitem{deng2009imagenet}
Jia Deng, Wei Dong, Richard Socher, Li{-}Jia Li, Kai Li, and Fei{-}Fei Li.
\newblock Imagenet: {A} large-scale hierarchical image database.
\newblock In {\em Proceedings of {IEEE/CVF} Conference on Computer Vision and
  Pattern Recognition}, pages 248--255, 2009.

\bibitem{Dhillon2020a}
Guneet~Singh Dhillon, Pratik Chaudhari, Avinash Ravichandran, and Stefano
  Soatto.
\newblock A baseline for few-shot image classification.
\newblock In {\em Proceedings of International Conference on Learning
  Representations}, 2020.

\bibitem{du2021smallgroup}
Xuefeng Du and Pengtao Xie.
\newblock Small-group learning, with application to neural architecture search,
  2021.

\bibitem{du2021learning}
Xuefeng Du, Haochen Zhang, and Pengtao Xie.
\newblock Learning by passing tests, with application to neural architecture
  search, 2021.

\bibitem{feurer2015initializing}
Matthias Feurer, Jost Springenberg, and Frank Hutter.
\newblock Initializing bayesian hyperparameter optimization via meta-learning.
\newblock In {\em Proceedings of the AAAI Conference on Artificial
  Intelligence}, volume~29, 2015.

\bibitem{finn2017modelagnostic}
Chelsea Finn, Pieter Abbeel, and Sergey Levine.
\newblock Model-agnostic meta-learning for fast adaptation of deep networks.
\newblock In {\em Proceedings of International Conference on Machine Learning},
  pages 1126--1135, 2017.

\bibitem{finn2017model}
Chelsea Finn, Pieter Abbeel, and Sergey Levine.
\newblock Model-agnostic meta-learning for fast adaptation of deep networks.
\newblock In {\em Proceedings of the 34th International Conference on Machine
  Learning-Volume 70}, pages 1126--1135. JMLR. org, 2017.

\bibitem{gidaris2019boosting}
Spyros Gidaris, Andrei Bursuc, Nikos Komodakis, Patrick P\'{e}rez, and Matthieu
  Cord.
\newblock Boosting few-shot visual learning with self-supervision.
\newblock In {\em Proceedings of IEEE/CVF International Conference on Computer
  Vision}, pages 8059--8068, 2019.

\bibitem{gidaris2018dynamic}
Spyros Gidaris and Nikos Komodakis.
\newblock Dynamic few-shot visual learning without forgetting.
\newblock In {\em Proceedings of IEEE/CVF Conference on Computer Vision and
  Pattern Recognition}, pages 4367--4375, 2018.

\bibitem{guillaumin2009is}
Matthieu Guillaumin, Jakob Verbeek, and Cordelia Schmid.
\newblock Is that you? metric learning approaches for face identification.
\newblock In {\em Proceedings of IEEE/CVF International Conference on Computer
  Vision}, pages 498--505, 2009.

\bibitem{frederick1978lbe}
Frederick Hayes-Roth.
\newblock Learning by example.
\newblock In {\em Cognitive Psychology and Instruction}, volume~5, pages
  27--38. Springer, Boston, MA, 1978.

\bibitem{frederick1976learning}
Frederick Hayes-Roth and John McDermott.
\newblock Learning structured patterns from examples.
\newblock In {\em Proceedings of the Third International Joint Conference on
  Pattern Recognition}, pages 419--423, 1976.

\bibitem{he2016resnet}
Kaiming He, Xiangyu Zhang, Shaoqing Ren, and Jian Sun.
\newblock Deep residual learning for image recognition.
\newblock In {\em Proceedings of IEEE/CVF Conference on Computer Vision and
  Pattern Recognition}, pages 770--778, 2016.

\bibitem{hosseini2021learning}
Ramtin Hosseini and Pengtao Xie.
\newblock Learning by self-explanation, with application to neural architecture
  search, 2021.

\bibitem{Hu2020Empirical}
Shell~Xu Hu, Pablo~Garcia Moreno, Yang Xiao, Xi~Shen, Guillaume Obozinski, Neil
  Lawrence, and Andreas Damianou.
\newblock Empirical bayes transductive meta-learning with synthetic gradients.
\newblock In {\em Proceedings of International Conference on Learning
  Representations}, 2020.

\bibitem{omid2021locality}
Omid Jafari, Preeti Maurya, Parth Nagarkar, Khandker~Mushfiqul Islam, and
  Chidambaram Crushev.
\newblock A survey on locality sensitive hashing algorithms and their
  applications.
\newblock {\em arXiv preprint arXiv:2102.08942}, 2021.

\bibitem{jang2017gumbel}
Eric Jang, Shixiang Gu, and Ben Poole.
\newblock Categorical reparameterization with gumbel-softmax.
\newblock In {\em Proceedings of the International Conference on Learning
  Representations}, 2017.

\bibitem{jiang2019learning}
Xiang Jiang, Mohammad Havaei, Farshid Varno, Gabriel Chartrand, Nicolas
  Chapados, and Stan Matwin.
\newblock Learning to learn with conditional class dependencies.
\newblock In {\em Proceedings of International Conference on Learning
  Representations}, 2019.

\bibitem{kar2012supervised}
Purushottam Kar and Prateek Jain.
\newblock Supervised learning with similarity functions.
\newblock In {\em Proceedings of Advances in Neural Information Processing
  Systems}, 2012.

\bibitem{kingma2015adam}
Diederik~P. Kingma and Jimmy Ba.
\newblock Adam: A method for stochastic optimization.
\newblock In {\em Proceedings of the International Conference on Learning
  Representations}, 2015.

\bibitem{koch2015siamese}
Gregory Koch, Richard Zemel, and Ruslan Salakhutdinov.
\newblock Siamese neural networks for one-shot image recognition.
\newblock In {\em Proceedings of International Conference on Machine Learning
  Deep Learning Workshop}, volume~2, 2015.

\bibitem{kostinger2012large}
Martin K\"{o}stinger, Martin Hirzer, Paul Wohlhart, Peter~M. Roth, and Horst
  Bischof.
\newblock Large scale metric learning from equivalence constraints.
\newblock In {\em Proceedings of IEEE/CVF Conference on Computer Vision and
  Pattern Recognition}, pages 2288--2295, 2012.

\bibitem{alex2012alexnet}
Alex Krizhevsky, Ilya Sutskever, and Geoffrey~E. Hinton.
\newblock {ImageNet} classification with deep convolutional neural networks.
\newblock In {\em Proceedings of Advances in Neural Information Processing
  Systems}, pages 1097--1105, 2012.

\bibitem{brenden2011one}
Brenden~M. Lake, Ruslan Salakhutdinov, Jason Gross, and Joshua~B. Tenenbaum.
\newblock One shot learning of simple visual concepts.
\newblock In {\em Cognitive Science Society}, 2011.

\bibitem{lee2019meta}
Kwonjoon Lee, Subhransu Maji, Avinash Ravichandran, and Stefano Soatto.
\newblock Meta-learning with differentiable convex optimization.
\newblock In {\em Proceedings of the IEEE/CVF Conference on Computer Vision and
  Pattern Recognition}, pages 10657--10665, 2019.

\bibitem{lee2018gradient}
Yoonho Lee and Seungjin Choi.
\newblock Gradient-based meta-learning with learned layerwise metric and
  subspace.
\newblock In {\em Proceedings of International Conference on Machine Learning},
  pages 2927--2936, 2018.

\bibitem{li2017metasgd}
Zhenguo Li, Fengwei Zhou, Fei Chen, and Hang Li.
\newblock {Meta-SGD}: Learning to learn quickly for few-shot learning.
\newblock {\em arXiv preprint arXiv:1707.09835}, 2017.

\bibitem{liu2018darts}
Hanxiao Liu, Karen Simonyan, and Yiming Yang.
\newblock {DARTS:} differentiable architecture search.
\newblock In {\em Proceedings of the International Conference on Learning
  Representations}, 2019.

\bibitem{liu2020prototype}
Jinlu Liu, Liang Song, and Yongqiang Qin.
\newblock Prototype rectification for few-shot learning.
\newblock In {\em Proceedings of European Conference on Computer Vision}, 2020.

\bibitem{liu2019learning}
Yanbin Liu, Juho Lee, Minseop Park, Saehoon Kim, Eunho Yang, Sungju Hwang, and
  Yi~Yang.
\newblock Learning to propagate labels: Transductive propagation network for
  few-shot learning.
\newblock In {\em Proceedings of International Conference on Learning
  Representations}, 2019.

\bibitem{brian2010metric}
Brian McFee and Gert~R. Lanckriet.
\newblock Metric learning to rank.
\newblock In {\em Proceedings of International Conference on Machine Learning},
  2010.

\bibitem{melekhov2016siamese}
Iaroslav Melekhov, Juho Kannala, and Esa Rahtu.
\newblock Siamese network features for image matching.
\newblock In {\em International Conference on Pattern Recognition}, pages
  378--383, 2016.

\bibitem{mishra2018a}
Nikhil Mishra, Mostafa Rohaninejad, Xi~Chen, and Pieter Abbeel.
\newblock A simple neural attentive meta-learner.
\newblock In {\em Proceedings of International Conference on Learning
  Representations}, 2018.

\bibitem{munkhdalai2018rapid}
Tsendsuren Munkhdalai, Xingdi Yuan, Soroush Mehri, and Adam Trischler.
\newblock Rapid adaptation with conditionally shifted neurons.
\newblock In {\em Proceedings of International Conference on Machine Learning},
  pages 3664--3673, 2018.

\bibitem{nguyen2019uncertainty}
Cuong Nguyen, Thanh-Toan Do, and Gustavo Carneiro.
\newblock Uncertainty in model-agnostic meta-learning using variational
  inference.
\newblock {\em arXiv preprint arXiv:1907.11864}, 2019.

\bibitem{nichol2018first}
Alex Nichol, Joshua Achiam, and John Schulman.
\newblock On first-order meta-learning algorithms.
\newblock {\em arXiv preprint arXiv:1803.02999}, 2018.

\bibitem{oreshkin2018tadam}
Boris~N. Oreshkin, Pau Rodriguez, and Alexandre Lacoste.
\newblock {TADAM}: Task dependent adaptive metric for improved few-shot
  learning.
\newblock In {\em Proceedings of Advances in Neural Information Processing
  Systems}, 2018.

\bibitem{qian2015similarity}
Qi~Qian, Inci~M. Baytas, Rong Jin, Anil Jain, and Shenghuo Zhu.
\newblock Similarity learning via adaptive regression and its application to
  image retrieval.
\newblock {\em arXiv preprint arXiv:1512.01728}, 2015.

\bibitem{qiao2018few}
Siyuan Qiao, Chenxi Liu, Wei Shen, and Alan Yuille.
\newblock Few-shot image recognition by predicting parameters from activations.
\newblock In {\em Proceedings of IEEE/CVF Conference on Computer Vision and
  Pattern Recognition}, pages 7229--7238, 2018.

\bibitem{ravi2017optimization}
Sachin Ravi and Hugo Larochelle.
\newblock Optimization as a model for few-shot learning.
\newblock In {\em Proceedings of International Conference on Learning
  Representations}, 2017.

\bibitem{rodriguez2020embedding}
Pau Rodr\'{i}guez, Issam Laradji, Alexandre Drouin, and Alexandre Lacoste.
\newblock Embedding propagation: Smoother manifold for few-shot classification.
\newblock In {\em Proceedings of European Conference on Computer Vision}, 2020.

\bibitem{rusu2019metalearning}
Andrei~A. Rusu, Dushyant Rao, Jakub Sygnowski, Oriol Vinyals, Razvan Pascanu,
  Simon Osindero, and Raia Hadsell.
\newblock Meta-learning with latent embedding optimization.
\newblock In {\em Proceedings of International Conference on Learning
  Representations}, 2019.

\bibitem{andrzej2012learning}
Andrzej Ruta and Yongmin Li.
\newblock Learning pairwise image similarities for multi-classification using
  kernel regression trees.
\newblock {\em Pattern Recognition}, 45(4):1396--1408, 2012.

\bibitem{adam2016metalearning}
Adam Santoro, Sergey Bartunov, Matthew Botvinick, Daan Wierstra, and Timothy
  Lillicrap.
\newblock Meta-learning with memory-augmented neural networks.
\newblock In {\em Proceedings of International Conference on Machine Learning},
  2016.

\bibitem{garcia2018fewshot}
Victor~Garcia Satorras and Joan~Bruna Estrach.
\newblock Few-shot learning with graph neural networks.
\newblock In {\em Proceedings of International Conference on Learning
  Representations}, 2018.

\bibitem{sheth2021learning}
Parth Sheth, Yueyu Jiang, and Pengtao Xie.
\newblock Learning by teaching, with application to neural architecture search,
  2021.

\bibitem{shu2019meta}
Jun Shu, Qi~Xie, Lixuan Yi, Qian Zhao, Sanping Zhou, Zongben Xu, and Deyu Meng.
\newblock Meta-weight-net: Learning an explicit mapping for sample weighting.
\newblock In {\em Advances in Neural Information Processing Systems}, pages
  1919--1930, 2019.

\bibitem{snell2017prototypical}
Jake Snell, Kevin Swersky, and Richard Zemel.
\newblock Prototypical networks for few-shot learning.
\newblock In {\em Proceedings of Advances in Neural Information Processing
  Systems}, 2017.

\bibitem{abs-1912-07768}
Felipe~Petroski Such, Aditya Rawal, Joel Lehman, Kenneth~O. Stanley, and Jeff
  Clune.
\newblock Generative teaching networks: Accelerating neural architecture search
  by learning to generate synthetic training data.
\newblock {\em CoRR}, abs/1912.07768, 2019.

\bibitem{sung2018learning}
Flood Sung, Yongxin Yang, Li~Zhang, Tao Xiang, Philip~H.S. Torr, and Timothy~M.
  Hospedales.
\newblock Learning to compare: Relation network for few-shot learning.
\newblock In {\em Proceedings of IEEE/CVF Conference on Computer Vision and
  Pattern Recognition}, pages 1199--1208, 2018.

\bibitem{triantafillou2017fewshot}
Eleni Triantafillou, Richard Zemel, and Raquel Urtasun.
\newblock Few-shot learning through an information retrieval lens.
\newblock In {\em Proceedings of Advances in Neural Information Processing
  Systems}, 2017.

\bibitem{vinyals2016matching}
Oriol Vinyals, Charles Blundell, Timothy Lillicrap, Koray Kavukcuoglu, and Daan
  Wierstra.
\newblock Matching networks for one shot learning.
\newblock In {\em Proceedings of Advances in Neural Information Processing
  Systems}, 2016.

\bibitem{wang2014learning}
Jiang Wang, Yang Song, Thomas Leung, Chuck Rosenberg, Jingbin Wang, James
  Philbin, Bo~Chen, and Ying Wu.
\newblock Learning fine-grained image similarity with deep ranking.
\newblock In {\em Proceedings of IEEE/CVF Conference on Computer Vision and
  Pattern Recognition}, pages 1386--1393, 2014.

\bibitem{wang2019simpleshot}
Yan Wang, Wei-Lun Chao, Kilian~Q. Weinberger, and Laurens van~der Maaten.
\newblock Simpleshot: Revisiting nearest-neighbor classification for few-shot
  learning.
\newblock {\em arXiv preprint arXiv:1911.04623}, 2019.

\bibitem{patrick1975learning}
Patrick~H. Winston.
\newblock Learning structural descriptions from examples.
\newblock In {\em Psychology of Computer Vision}. McGraw-Hill, New York, 1975.

\bibitem{wu2017tiny}
Jiayu Wu, Qixiang Zhang, and Guoxi Xu.
\newblock Tiny imagenet challenge.
\newblock In {\em cs231n}, Technical Report, 2017.

\bibitem{xie2021skillearn}
Pengtao Xie, Xuefeng Du, and Hao Ban.
\newblock Skillearn: Machine learning inspired by humans' learning skills,
  2021.

\bibitem{ye2020fewshot}
Han-Jia Ye, Hexiang Hu, De-Chuan Zhan, and Fei Sha.
\newblock Few-shot learning via embedding adaptation with set-to-set functions.
\newblock In {\em Proceedings of IEEE/CVF Conference on Computer Vision and
  Pattern Recognition}, pages 8808--8817, 2020.

\bibitem{zhao2021learning}
Xingchen Zhao, Xuehai He, and Pengtao Xie.
\newblock Learning by ignoring, with application to domain adaptation, 2021.

\bibitem{zhengmeta19}
Guoqing Zheng, Ahmed~Hassan Awadallah, and Susan~T. Dumais.
\newblock Meta label correction for learning with weak supervision.
\newblock {\em CoRR}, abs/1911.03809, 2019.

\end{thebibliography}

% \newpage
%%%%%%%%%%%%%%%%%%%%%%%%%%%%%%%%%%%%%%%%%%%%%%%%%%%%%%%%%%%%
% \input{SECTIONS/70_Checklist/manuscript}

% \newpage
% %%%%%%%%%%%%%%%%%%%%%%%%%%%%%%%%%%%%%%%%%%%%%%%%%%%%%%%%%%%%
% \input{SECTIONS/Appendix/manuscript}

\end{document}